\relax
\documentclass[letterpaper]{article} % DO NOT CHANGE THIS
\usepackage{aaai22}  % DO NOT CHANGE THIS
\usepackage{times}  % DO NOT CHANGE THIS
\usepackage{helvet}  % DO NOT CHANGE THIS
\usepackage{courier}  % DO NOT CHANGE THIS
\usepackage[hyphens]{url}  % DO NOT CHANGE THIS
\usepackage{graphicx} % DO NOT CHANGE THIS
\usepackage{natbib}  % DO NOT CHANGE THIS AND DO NOT ADD ANY OPTIONS TO IT
\usepackage{caption} % DO NOT CHANGE THIS AND DO NOT ADD ANY OPTIONS TO IT
\DeclareCaptionStyle{ruled}{labelfont=normalfont,labelsep=colon,strut=off} % DO NOT CHANGE THIS
\usepackage{algorithm}
\usepackage{algorithmic}

\usepackage{graphicx}
\usepackage{amsmath}
\usepackage{amssymb}
\usepackage{array}

\usepackage{float}  %设置图片浮动位置的宏包
\usepackage{multirow}
\usepackage{algorithm}
\usepackage{algorithmic}
\usepackage{caption}
\usepackage{epsfig}
\usepackage{amsmath, amsthm, amsfonts, amssymb, gensymb}
\usepackage{booktabs}
\usepackage{appendix}
\usepackage{makecell}
\usepackage{subfloat}
\usepackage{subcaption}
\usepackage{setspace}
\usepackage{textcomp}
\usepackage{gensymb}
\usepackage{verbatimbox}

\newcommand{\ie}{\emph{i.e.}}

\usepackage{amsfonts}
\usepackage{amssymb}
\usepackage{amsthm,amsmath}
\usepackage{mathrsfs}
\usepackage{indentfirst}
\usepackage{multirow}

\usepackage{color}

\newcommand{\red}[1]{{\color{red}{#1}}}

\usepackage[pagebackref,breaklinks,colorlinks]{hyperref}

\usepackage[capitalize]{cleveref}
\crefname{section}{Sec.}{Secs.}
\Crefname{section}{Section}{Sections}
\Crefname{table}{Table}{Tables}
\crefname{table}{Tab.}{Tabs.}
\usepackage{newfloat}
\usepackage{listings}
\floatstyle{ruled}
\newfloat{listing}{tb}{lst}{}
\floatname{listing}{Listing}

\title{Absolute Zero-Shot Learning}
\author {
   Rui Gao\equalcontrib \textsuperscript{\rm 1},
Fan Wan\equalcontrib \textsuperscript{\rm 2},
Daniel Organisciak\textsuperscript{\rm 3},
 Jiyao Pu\textsuperscript{\rm 2},\\ 
   Junyan Wang\textsuperscript{\rm 4},
    Haoran Duan\textsuperscript{\rm 2},
    Peng Zhang\textsuperscript{\rm 2},
    Xingsong Hou\textsuperscript{\rm 1},
    Yang Long\textsuperscript{\rm 2}
}

\affiliations {
    \textsuperscript{\rm 1} Xi’an Jiaotong University, Xi'an, China\\
    \textsuperscript{\rm 2} Durham University, Durham, UK\\
    \textsuperscript{\rm 3} Northumbria University, Newcastle, UK\\
    \textsuperscript{\rm 4} The University of New South Wales, Sydney, Australia\\
    gaorui1013@stu.xjtu.edu.cn,
    \{fan.wan,jiyao.pu,haoran.duan,peng.zhang\}@durham.ac.uk,
    daniel.organisciak@northumbria.ac.uk,
    z5307612@ad.unsw.edu.au,
    houxs@mail.xjtu.edu.cn,
    yang.long@ieee.org
}

\usepackage{bibentry}
\begin{document}

\maketitle

%%%%%%%%% ABSTRACT
\begin{abstract}
 %Zero-Shot Learning (ZSL) has provided a promising solution to generalise a pre-trained model to recognise unseen classes. However, the training process relies on access to a large proportion of data. 
 Considering the increasing concerns about data copyright and privacy issues, we present a novel Absolute Zero-Shot Learning (AZSL) paradigm, \ie, training a classifier with zero real data. The key innovation is to involve a teacher model as the data safeguard to guide the AZSL model training without data leaking. The AZSL model consists of a generator and student network, which can achieve date-free knowledge transfer while maintaining the performance of the teacher network. We investigate `black-box' and `white-box' scenarios in AZSL task as different levels of model security. Besides, we also provide discussion of teacher model in both inductive and transductive settings. Despite embarrassingly simple implementations and data-missing disadvantages, our AZSL framework can retain state-of-the-art ZSL and GZSL performance under the `white-box’ scenario. Extensive qualitative and quantitative analysis also demonstrates promising results when deploying the model under 'black-box' scenario.
\end{abstract}

%%%%%%%%% BODY TEXT
\section{Introduction}\label{sec:Intro}
The blossom of deep learning technologies embraces the development of high-performance computing and large-scale multi-modal data. The nature of deep learning is that the pre-trained base model can extract empirical knowledge, \textit{e.g.} visual features \cite{he2016deep} and semantic meanings \cite{devlin2018bert} from large-scale datasets. However, sharing data across different institutes and even between different countries has become increasingly difficult and sensitive. The increasing awareness of data copyright, expensive data cleaning and annotation, and restricted access to data in expert domains, e.g. health and security, has hindered the development of interdisciplinary and intercultural deep models.

\begin{figure}[t]
    \centering
    \includegraphics[width=0.9\linewidth]{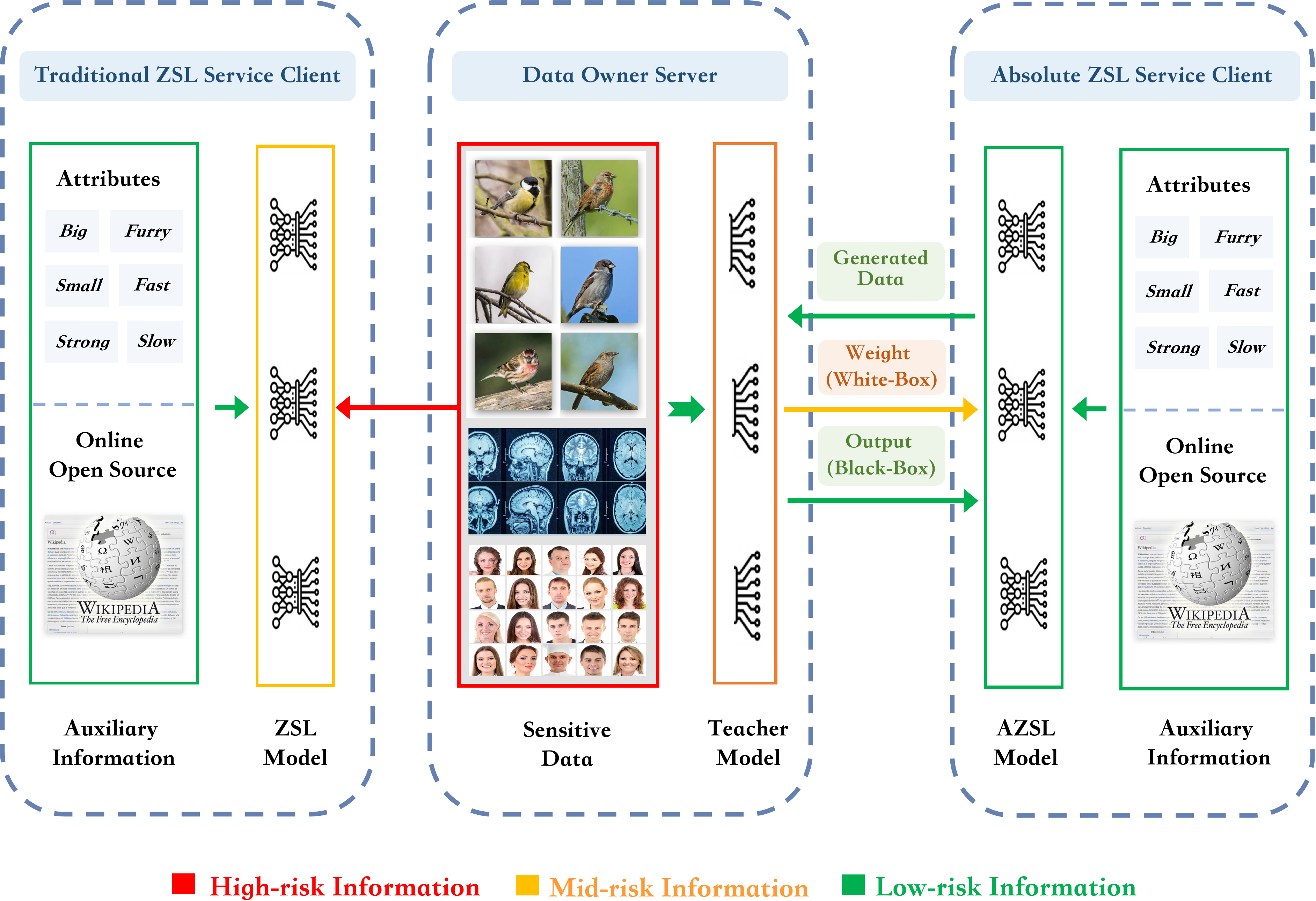}
    \caption{Traditional ZSL models require access to real images from the data owner to learn the visual-semantic associations. AZSL suggests an extra data safeguard using a teacher model so that an AZSL model can achieve GZSL without access to any real images. The training of AZSL only involves generated data and prior auxiliary information and guidance from the teacher model.}
    \label{zsl-azsl}
\end{figure}

As shown in Fig.\ref{zsl-azsl}, datasets may contain biometric, health, face information etc. \cite{ukbiobank} that cost the data owner over billions and tens of years to collect. Strict regulations, such as the GDPR\cite{voigt2017eu} in Europe has been enforced to control the risk of data leaking. When an AI company needs access to such sensitive data to provide AI services, the shared data is confronting the leaking risk even though tedious confidentiality agreements have been signed. 

This paper is motivated by the increasing concerns about data and model privacy issues. As a promising machine learning paradigm, Zero-Shot Learning (ZSL) investigates an extreme case when such deep transfer cannot be achieved due to no access to the test domain data \cite{lampert2013attribute,fu2015transductive,annadani2018preserving}.
%The initial stage of ZSL evolution (2008-2017) concentrates on model development. 
The learning objective of ZSL is to build a consistent visual-semantic mapping based on auxiliary information (\ie, attributes or word embeddings) on seen class domain $S$ so that model can generalise to unseen class domain $U$ at the test time \cite{lampert2009learning,frome2013devise}. %Such settings have received considerable attention in many areas, such as \cite{xian2017zero}, which indicates the sign of entering the second ZSL evolution in applications.
The criticizing and rethinking in ZSL task never ceases. For available data during training, existing settings can be divided into inductive and transductive ZSL. Early ZSL models focus mainly on inductive ZSL (IZSL) \cite{lampert2009learning,jayaraman2014zero}, which only utilise labeled seen data for training. Transductive ZSL (TZSL)  \cite{kodirov2015unsupervised,song2018transductive} assumes both seen labeled and unseen unlabeled data are available during the training stage, which aims to utilise the universe information for improving the generalization of model. Besides, generalization ability is also a key issue for ZSL task. In test phase, early ZSL models \cite{norouzi2013zero,lampert2013attribute} can only output unseen class labels during evaluation, which is called conventional ZSL (CZSL), \ie, test images only come from unseen classes. Generalised ZSL (GZSL) \cite{chao2016empirical} is then proposed, which aims to combine seen and unseen classifiers into a unified model.

Existing ZSL models are established based on real data from either seen or unseen classes. When adapting a pre-trained model to a new task domain, existing ZSL models assume a large amount of labeled seen class or unlabeled unseen class data are available to establish the visual-semantic relationship. However, sharing data across different institutes and even between different countries is often infeasible. Different from existing ZSL settings, we focus on establishing ZSL model without data sharing during the training process.  In this paper, we propose a new paradigm dubbed Absolute Zero-Shot Learning (AZSL) to avoid sensitive data leaking while still enabling AI model can be trained. Figure \ref{zsl-azsl} briefly illustrates the difference between ZSL and AZSL task. As for data privacy preserving, federated learning \cite{mcmahan2017communication,hao2021towards} provides a promising strategy through model sharing between data owner server and service client. Motivated by this, our AZSL task suggests replacing data with a teacher model (pre-trained on real data) to guide the ZSL model training. Teacher model can be regarded as the implicit representation of data so that AZSL model can be established through the supervision of teacher model, which can prevent real data from being shared. While federated learning has demonstrated promise for data privacy, it only tackles the supervised learning task, which cannot be extended to unseen class classification. Here we propose the data-free knowledge transfer framework for AZSL task. Concretely, we aim to synthesise data conditioned on class-level semantic embeddings depending on the guidance of teacher model, so that ZSL classification will be possible based on synthesised data. 
 
To comprehensively explore our proposed AZSL framework, we also present extensive discussion from the perspective of model weights security and knowledge space of the teacher model. First, we propose two AZSL scenarios in terms of model security. When data owner provides the teacher model as data safeguard for ZSL model training, the data may be leaked to some extent by gradients exchange \cite{zhu2019deep}. To explore the influence of different guidance provided by the teacher model, we propose the AZSL model in two scenarios for further discussion. In the `black-box' scenario, the teacher only provides output classification scores but does not share weights. In the `white-box' scenario, the teacher will also share the model weights during training, which is more informative. These two scenarios indicate different levels of communication between data owner and AI service provider, which will lead to different ZSL recognition performance. Furthermore, we propose two types of teachers for discussion. Existing ZSL methods can be categorised into IZSL and TZSL depending on whether the unlabeled unseen data are available for training. Motivated by this, we propose inductive and transductive teachers according to whether unseen classes are involved in pre-training teacher model. It is worth discussing the generalization ability of AZSL model with different knowledge spaces when training teacher model. In summary, our contributions are three-fold: 
\begin{itemize}
\setlength{\itemsep}{0.5pt}
\setlength{\parsep}{0.5pt}
\setlength{\parskip}{0.5pt}
    \item Absolute Zero-Shot Learning aims to achieve zero-shot classification without access to real data. The paradigm can be applied to many real-world applications when access to real data is not permitted.
    \item We develop a novel data-free knowledge transfer framework for AZSL task. In addition to the zero data sharing setting, we propose `black-box' and `white-box' scenarios and discuss the pros and cons of model sharing problems. We also present analysis of teacher model in both inductive and transductive settings.
    \item We show experimental results for both conventional ZSL and GZSL tasks in two scenarios. Our AZSL model achieves promising performance on existing benchmarks despite the disadvantage of data absence. Extensive qualitative analysis demonstrates the effectiveness of our framework. 
\end{itemize}

\renewcommand\arraystretch{5}

\begin{table*}[t]
\small
\caption{Differences between AZSL and existing ZSL settings. `S' and `U' represent seen and unseen class. `$\mathcal X$' represents visual features. `$\tilde{\mathcal{X}}$' represents generated features. `$\theta$' represents the ZSL model and `$\theta_T$' represents the pre-trained teacher model in AZSL task.}
\resizebox{0.98\textwidth}{!}
{

\begin{tabular}{ccp{4.8cm}<{\centering}cc}
\hline\hline\\[-70pt]
                   %& Paradigm    & Leaking Data & Leaking Weights 
                    & IZSL    & TZSL& AZSL \\[-10pt]\hline

\specialrule{0em}{-10pt}{-15pt}

\makecell[c]{Leaking Data}              &  $\mathcal X_s$       & $\mathcal X_s \cup \mathcal X_u$   &  \textbf{0}   \\

\specialrule{0em}{-15pt}{-18pt}

\makecell[c]{Leaking Weights}   &  $\theta_S$    &    $\theta_{S+U}$      & \textbf{$\theta_{T}$} $/$ \textbf{0}   \\

\specialrule{0em}{8pt}{-15pt}

\makecell[c]{Paradigm}                &    \begin{minipage}[b]{0.5\columnwidth}
		\centering
		\raisebox{-.5\height}{\includegraphics[width=\linewidth]{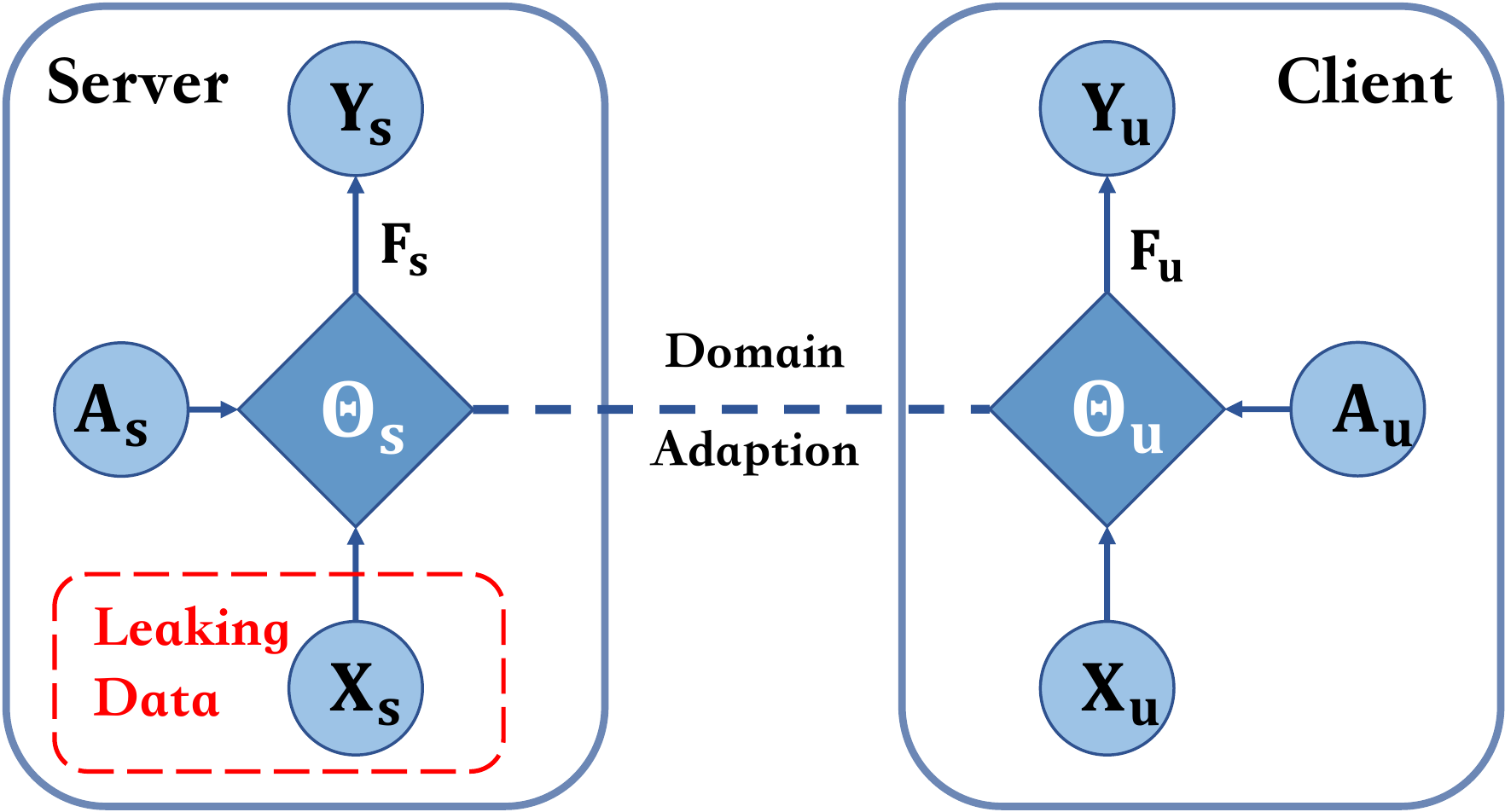}}
	\end{minipage}     &  \begin{minipage}[b]{0.5\columnwidth}
		\centering
		\raisebox{-.5\height}{\includegraphics[width=\linewidth]{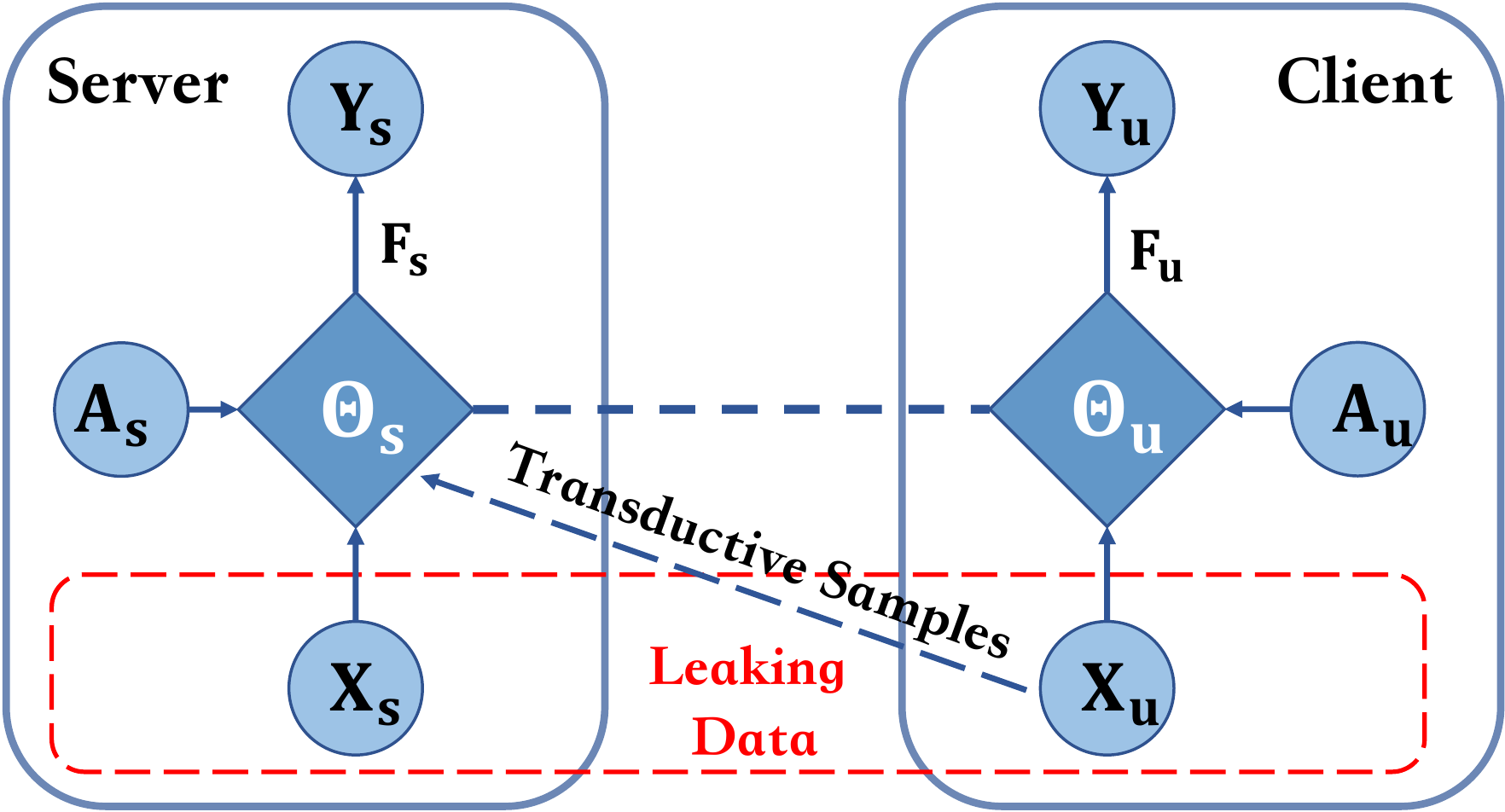}}
	\end{minipage} &   \begin{minipage}[b]{0.5\columnwidth}
		\centering
		\raisebox{-.5\height}{\includegraphics[width=\linewidth]{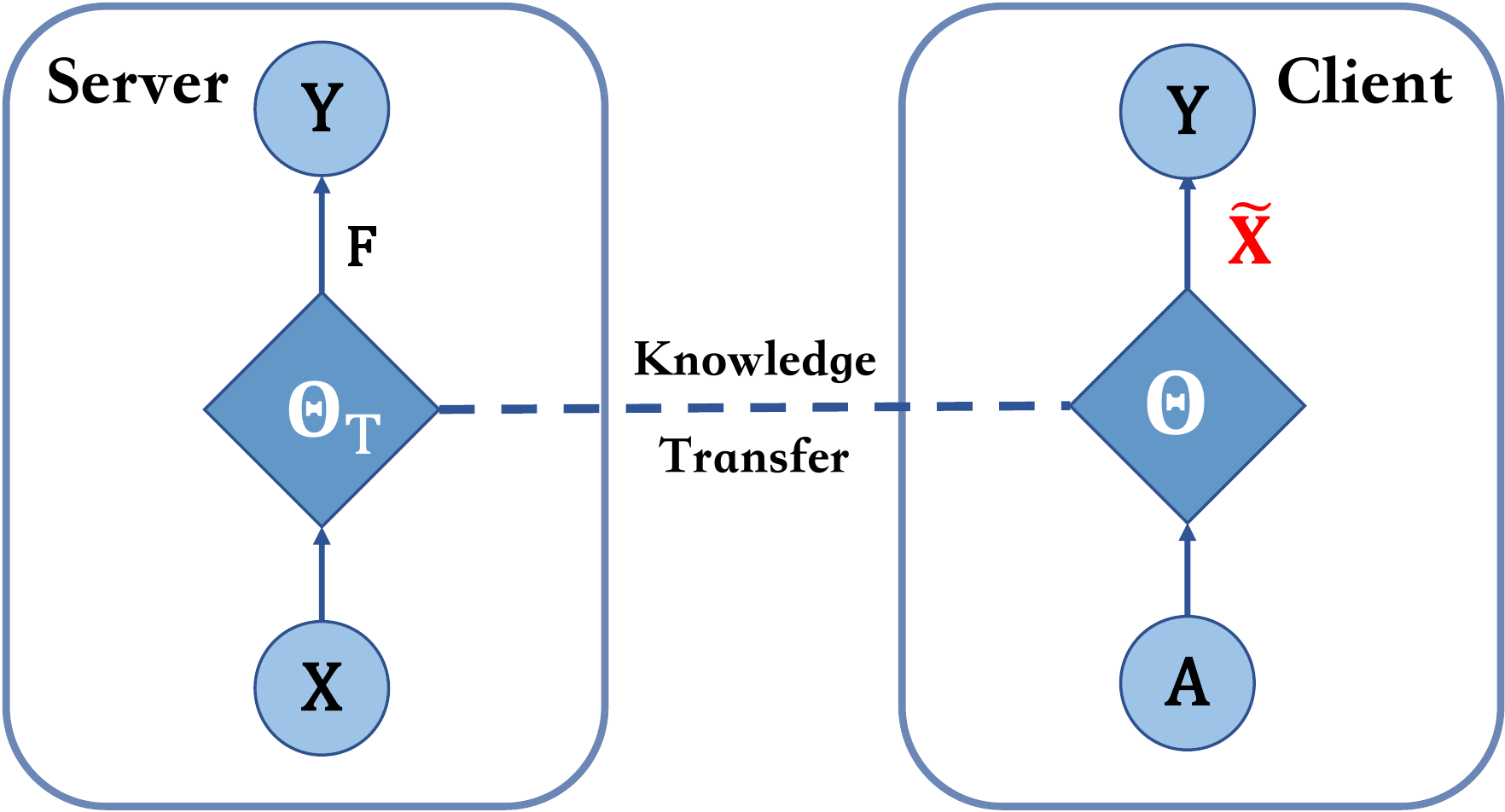}}
	\end{minipage}     \\

\specialrule{0em}{5pt}{0pt}
\hline\hline                      
\end{tabular}\label{formulation} 
}
\end{table*}

\renewcommand\arraystretch{1}

\section{Related Work}
The most widely used framework for data privacy preserving is the federated learning \cite{konevcny2016federated, mcmahan2017communication}. A global model is shared with clients to avoid data leaking. However, data-free knowledge transfer and model inversion techniques \cite{yin2020dreaming,smith2021always,yin2021see} can recover data from the pre-trained model and thus can be used to attack shared models in federated learning. Knowledge distillation utilises the domain-expert teacher model to train a compact student model with pursuing competitive recognition accuracy \cite{xu2020computation,chen2021wasserstein}. Machine teaching \cite{zhu2015machine,liu2017iterative} focuses on training a teacher model to select an optimal data set for the student model. Both knowledge distillation and machine teaching provides promising solutions to separate student model so that teacher model can avoid the attack. Yet, none of these methods have explored their potential in data privacy preserving. This paper presents the first work exploring the data privacy potential via zero-shot learning paradigm. Meanwhile, our proposed AZSL retains the property of traditional ZSL that can generalised to unseen classes without further training.

Zero-Shot learning \cite{zhang2018triple,larochelle2008zero,vyas2020leveraging,ma2020variational} aims to recognise unknown/unseen classes by establishing the relationship between seen and unseen classes, which can bridge the gap via class semantic information, \ie, attributes \cite{jayaraman2014zero}, predefined similes \cite{long2017describing} and word embedding \cite{zhang2016zero}. In terms of establishing the relationship between visual and semantic spaces, existing methods can be mainly separated into three groups.
Some works \cite{qin2016beyond,kodirov2017semantic,felix2018multi} aim to build the mapping from visual to semantic space and then assign class label via nearest neighbor (NN) classifier. Other works \cite{gao2020zero,xian2018feature} focus on unseen class data generation to alleviate data-missing problem. Meanwhile, \cite{akata2013label,akata2015evaluation} explore an effective common space for visual and semantic embedding by maximizing their compatibility scores. existing ZSL methods aim to build the effective mapping among visual and semantic spaces to mitigate the gap between seen and unseen classes.

According to whether unseen data is adopted during training, existing ZSL methods can be categorised into inductive \cite{long2017zero,romera2015embarrassingly} and transductive settings \cite{song2018transductive,fu2015transductive}. As for the test phase, inductive ZSL methods \cite{akata2013label,norouzi2013zero} assume the test data only come from unseen classes. Generalised ZSL (GZSL) \cite{chao2016empirical,min2020domain} is then proposed to assign both seen and unseen data into corresponding classes. Compared with conventional ZSL, GZSL attempts to model the more challenging real situation. 

%There has been little research on zero-shot learning whilst maintaining data privacy. So we propose a new AZSL setting, which aims to accomplish the zero-shot recognition without access to real data during training. 
We present the differences between AZSL and existing ZSL settings in Table \ref{formulation}. In terms of leaking data in the training process, only labeled seen data is available in IZSL while TZSL can use data from both seen and unseen classes. AZSL setting assumes that model is established without access to any real data. As for model security, we refer to the sharing of weights trained on real data as weight leaking. ZSL model weights in both inductive and transductive settings are all leaked. For AZSL, teacher model pre-trained on real data is provided during training. In terms of teacher weights privacy, we propose black- and white-box scenarios. The teacher weights are leaked for guidance during training AZSL models in white-box scenario while no weights are shared in black-box scenario, which preserves the privacy for both data and model weights.

\begin{figure*}
    \centering
    \includegraphics[width=0.7\linewidth]{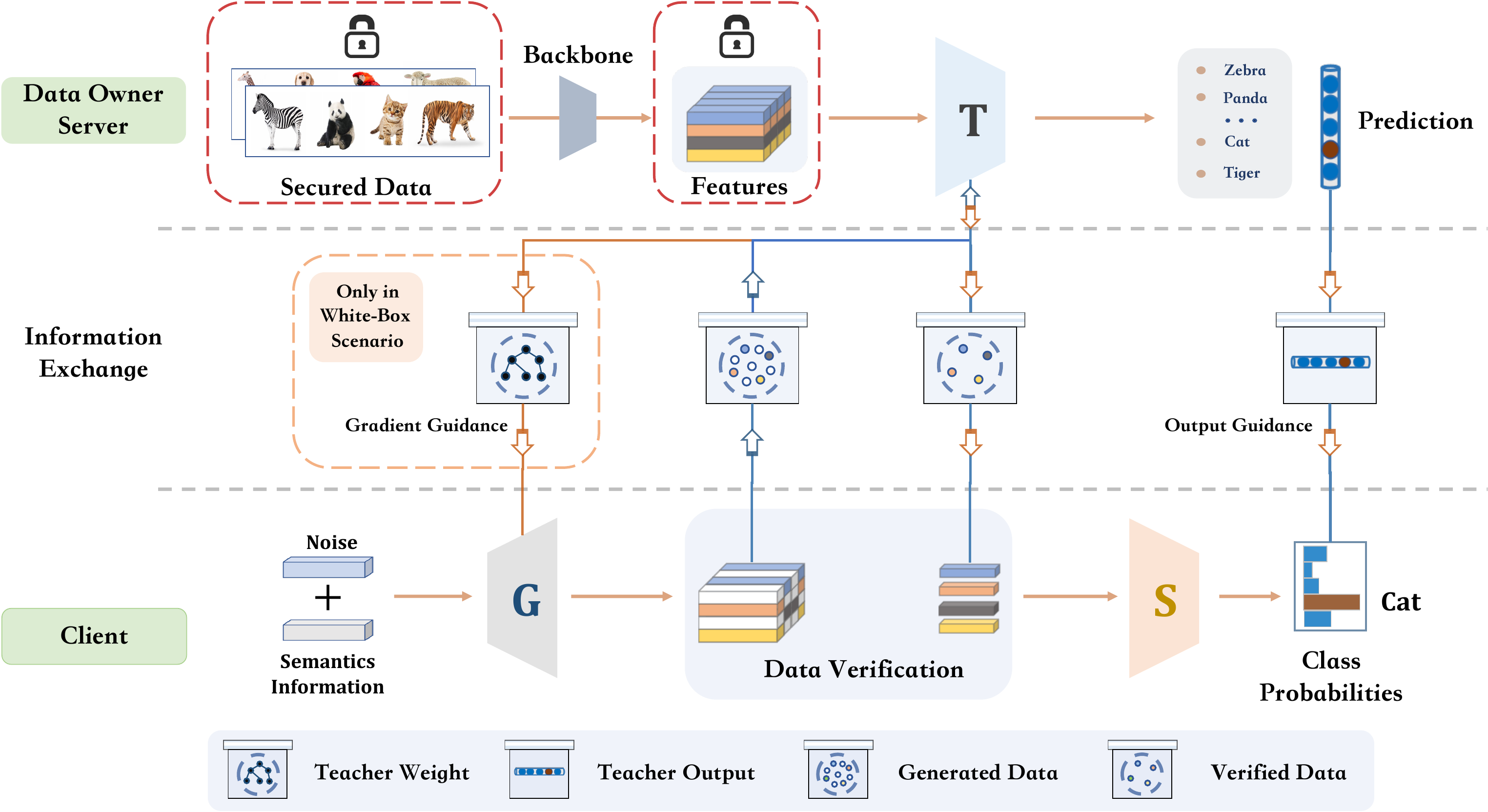}
    \caption{Overall framework in the black-box and white-box scenarios. Our framework can generate high-quality features with the guidance of teacher. In white-box scenario, the generator has access to teacher weights during training while teacher only provides output guidance in black-box scenario. }
    \label{AZSL-framework}
\end{figure*}

\section{Absolute Zero-Shot Learning}
As shown in Fig. \ref{AZSL-framework}, AZSL addresses the problem when sensitive data is secured on the \textit{Server} domain. The key idea is to introduce a teacher model as the data safeguard and guide the model deployed on the \textit{Client} domain to train an AZSL classifier with absolute zero real data. In addition to data privacy preserving, we introduce white-box and black-box scenarios to discuss the teacher model sharing problem regarding the balance between performance and security.

\subsection{AZSL Problem Formulation}
The basic AZSL setting involves secured images and their extracted visual features $x\in\mathcal{X}$. The data safeguard is a pre-trained teacher model on the server domain. For simplicity, we consider the supervised learning model $f_{T}: \mathcal{X}\rightarrow\mathcal{Y}$, where $y\in\mathcal{Y}$ is the label space. The ultimate goal of AZSL is to train a student model on the client domain using an objective function that learns from the guidance of the teacher model: 
\begin{equation}
\setlength{\abovedisplayskip}{5pt} 
\setlength{\belowdisplayskip}{5pt}
    \ell\left( f_{AZSL}\left(\tilde{x}\right), f_{T}\left(\tilde{x}\right) \right), 
    \label{E1}
\end{equation}
where $\tilde{x}\in\tilde{\mathcal{X}}$ is generated data that ensures no real data can be accessed.

%Different from ZSL training, AZSL aims to accomplish zero-shot classification without access to real data in both seen and unseen classes during model training. The core idea is to introduce a pre-trained teacher model that provides guidance and takes the place of real data to train AZSL framework. Real data is held by the data owner server, which cannot be shared to the AI service client. The teacher is trained on real data to act as data safeguard, which is provided to the service client during AZSL model training. 

\noindent\textbf{Inductive vs Transductive AZSL}
On the server domain, we further breakdown the AZSL into inductive (Ind) and transductive (Trans) teachers according to the label space. Seen classes are defined as $\mathcal S = \{ \left ( x_s, a_s, y_s \right ) \mid$ $ x_s \in \mathcal X_s, a_s \in \mathcal A, y_s \in \mathcal Y_s \}$, where $x_s \in \mathbb{R}^{d_x}$ denotes the  $d_x$-dimensional visual feature in the set of seen class features, $a_s \in \mathbb{R}^{d_a}$ denotes the $d_a$-dimensional auxiliary class-level semantic embedding, and $\mathcal Y_s$ stands for the set of labels for seen classes. Unseen classes are defined as $\mathcal U = \{ \left ( x_u,a_u,y_u \right ) \mid$ $ x_u \in \mathcal X_u, a_u \in \mathcal A, y_u \in \mathcal Y_u\}$, where $x_u$ represents the unseen class features, $a_u$ denotes the class-level semantic embedding of unseen classes and $y_u$ denotes the unseen class labels. 
Seen and unseen classes are disjoint, \ie, $\mathcal Y_s \cap \mathcal Y_u = \emptyset$. For AZSL, both seen and unseen features, $x_s$ and $x_u$, are unavailable for client. Available information for AZSL client can be represented as $\mathcal T_r= \{ \left ( a, y \right ) \mid a \in \mathcal A, y \in \mathcal Y \}$, which means only semantic embedding and class labels can be accessed. In this way, we consider the basic AZSL $f_{T}: \mathcal{X}\rightarrow\mathcal{Y}$ is transductive because the source domain contains both seen and unseen classes. A more challenging Inductive AZSL considers  $f_{T}: \mathcal{X}_s\rightarrow\mathcal{Y}_s$.

\noindent\textbf{ZSL vs GZSL} On the client domain, AZSL aims classify test images $f_{AZSL}:\mathcal X_{u} \rightarrow \mathcal Y_u$ for conventional ZSL, and $f_{GAZSL}:\mathcal X \rightarrow \mathcal Y$ for Generalised ZSL. The training of above classifiers using absolutely generated data will be introduced next.

\subsection{White-Box \& Black-Box Scenarios}
The objective function of AZSL in Eq. \ref{E1} defines a data-free knowledge transfer framework for AZSL task, \ie, through the guidance of the teacher model, our proposed AZSL formula consists of two components: a \textit{Generator} $G$ and a \textit{Student} network $S$.
Figure \ref{AZSL-framework} depicts the detailed AZSL framework in both white-box and black-box scenarios. The system consists of 1) the secured data and teacher model on the server; 2) AZSL model on the client; 3) and the information exchange channels. %Synthesised data by $G$ is packed as a request to upload to the server. The teacher model receives the request data and provides feedback to guide the training of $G$. $S$ is then trained with synthesised data to approach the classification performance of teacher model without access to the real data.

Considering the model inversion can attack share models, we also investigate the security levels regarding model sharing in addition to data privacy preserving. For white-box scenario, the teacher weights are involved to compute the gradient for the generator and student network training. In the black-box scenario, the teacher only provides the output as pseudo labels, \ie, teacher model is not involved in back propagation for optimization of AZSL framework.

\paragraph{White-Box Scenario}
In white-box scenario, the teacher model provides both gradient and softmax output as the AZSL training guidance as follows. \textbf{1) Uploading generated data:} the generator synthesises features of same classes with the supervision with teacher based on noise $z$ and class-level semantic embedding $a$ (attributes or BERT model of class names \cite{devlin2018bert} as the condition). Specifically, we aim to synthesise the features that can be classified into corresponding classes with the constraint of the teacher network. $\tilde{x} = G(z|a; \theta_{G})$ represents the generated features which are uploaded to the server. \textbf{2) Gradient and softmax guidance:} The teacher model receives $\tilde{x}$ and process the data using the loss function:
\iffalse
\begin{equation}\label{5}
    \min_{\theta_{G}}  -\mathbb{E}_{z \sim p_z}\left[ \log \left(T^*\left(G(z|a; \theta_{G}); \theta_{T}^*\right)\right) \right],
\end{equation}
\fi
\begin{equation}\label{5}
\setlength{\abovedisplayskip}{5pt} 
\setlength{\belowdisplayskip}{5pt}
    \min_{\theta_{G}}  \mathcal L(\tilde{x},y;\theta_G)+ \alpha \mathcal R(\tilde{x}),
\end{equation}
where $\mathcal L(\cdot)$ represents cross-entropy loss by teacher model for classification, $\mathcal{R}(\cdot)$ refers to the regularisation term during feature generation with hyperparameter $\alpha$ . The regularisation term aims to minimise the distribution distance of real and generated features. Note that the regularization is also completed at the server side and real data will not be accessed for the client. \textbf{3) Feedback downloading}: a request is sent to the client so that the gradient, the regularisation of distribution divergence, and softmax output can be download. \textbf{4) Label verification:} Using the softmax to compute pseudo labels and filter out misclassified generated samples:
\begin{equation}\label{3}
\setlength{\abovedisplayskip}{5pt} 
\setlength{\belowdisplayskip}{5pt}
\begin{aligned}
(\tilde{x}^{*},y^{*}) \in \{&(\tilde{x}, y)|y=\mathop{argmax}T(\tilde{x};\theta _{T}^*),\\
&\tilde{x}=G(z|a;\theta _{G}^*)\},
\end{aligned}
\end{equation}
where $\tilde{x}^{*}$ denotes the high-quality generated features and $y^{*}$ denotes the corresponding class labels. \textbf{5): Training the student network:}
\begin{equation}\label{4}
\setlength{\abovedisplayskip}{5pt} 
\setlength{\belowdisplayskip}{5pt}
    \min_{\theta_{S}}  \left \| T^{*} (\tilde{x}^{*};\theta_{T}^{*})-S(\tilde{x}^{*};\theta_{S})\right \|_{2}^{2}.
\end{equation}
%The student network is optimised to imitate the behavior of the teacher by matching the softmax output of the teacher network which represents the real class distribution. It is desirable to generate features that can be correctly classified by the student network.   
In the white-box scenario, the gradient is imposed directly onto the generated features and can massively improve the performance of the generator. As a trade-off, the gradient feedback is mid-risk information (may lead to teacher model leaking) whereas the softmax and regularisation feedback are low-risk. 

\begin{algorithm}[t]
    \setstretch{0.62}
    \small
    \caption{Training Procedure in both Scenarios}     
    \label{white-black}
    \begin{algorithmic}[1] 
    \REQUIRE ~~\\  
     Pre-trained Teacher network $\theta_{T}^*$, class labels $\mathcal Y_{tr}$ and their auxiliary semantic embedding $\mathcal{A}$; the maximal number of training epochs $T_g$ and $T_s$ for generator and student network, respectively.
    \ENSURE ~~\\  
    The learned parameters $\theta_{G} $, $ \theta_{S}$ for generator $G$ and student network $S$, respectively.
    
    \STATE Initializing, $\theta _{G} $, $ \theta _{S}$. Set the iteration epoch $t_g=1,t_s=1$
    
    \WHILE{ $t_g<T_g$ }  
    \IF{White-Box Scenario}
    \STATE Training generator with the gradient guidance of the teacher network through Eq.(\ref{5}).
    \ELSIF{Black-Box Scenario}
    \STATE Training generator with the output guidance of the teacher network through Eq.(\ref{2}).
    
    \ENDIF
    \STATE $t_g:=t_g+1$;
    \ENDWHILE  
    \STATE Conducting data verification through Eq.(\ref{3}).
    
    \WHILE{ $t_s<T_s$ }  
    \STATE Training student network with the output guidance of teacher network through Eq.(\ref{4}).
    \STATE $t_s:=t_s+1$;
    \ENDWHILE   
    \end{algorithmic}  
\end{algorithm}

\paragraph{Black-Box Scenario}
Black-box scenario only differs from the white-box scenario in the guidance provided by the teacher model in the second step. Only low-risk regularisation and softmax output can be requested from the teacher model so as to avoid the model leaking risk.
%any real data or model is not leaked during AZSL framework training, \ie, the teacher weights are also unavailable and they cannot be involved in the back propagation optimization process. Compared with white-box scenario, we modify the stage of generator training, \ie, the teacher network only provides the output guidance during the training process of the generator and student model.  
%To train the generator, the student network is involved due to the inaccessibility of teacher model weights. 
Specifically, the generated features $\tilde{x} = G(z|a; \theta_{G})$ is uploaded to the server to compute the softmax and divergence regularisation. The server then creates a request so that the feedback can be downloaded. Generated data can validate whether its conditional class input can match the teacher softmax output and misclassified samples are filtered out. The generator $G$ and student network $S$ are then trained as an end-to-end model as follows:
\begin{equation}\label{2}
\setlength{\abovedisplayskip}{6pt} 
\setlength{\belowdisplayskip}{6pt}
    \min_{\theta_{G},\theta_{S}}  \left \| T^{*} (\tilde{x};\theta_{T}^{*})-S(\tilde{x};\theta_{S})\right \|_{2}^{2}+\alpha \mathcal{R}(\tilde{x}),
\end{equation}
where $\theta_{T}^{*}$ denotes the optimised parameters of teacher, $\theta_{G},\theta_{S}$ denotes the parameters of generator and student network respectively. %Through the training with the output guidance of the teacher network, the generator has the ability to synthesise features of corresponding classes. 

In summary, there are two crucial tasks in AZSL, \textit{i.e.} data generation and verification. As the first AZSL, our key contributions focus on investigating key research questions: 1) what are the impacts of different teacher feedback information on the quality and diversity of generated data? 2) different semantic information as generation condition and their impacts; 3) the trade-off between performance and security in white-box and black-box; 4) can student generate new knowledge beyond the limitation of an inductive teacher?
%Similar to the white-box scenario, data verification is then adopted for selection of generated features and student network is fine-tuned on filtered generated features with the output guidance of teacher model. With the supervision of teacher, the student network can achieve data-free knowledge transfer for AZSL classification task. 
5) previous work uses real seen data and generated unseen data, which causes the prediction bias towards seen classes. In AZSL, both seen and unseen classes are trained using generated data, which improves the consistency between seen and unseen classifier in the GZSL problem. Detailed training procedure in both scenarios is shown in Algorithm \ref{white-black}.

\subsection{Absolute Zero-Shot Classification}

After the training process, the generator can synthesise features of good quality and the student network can predict class labels of test features. %Recall that we train the teacher with transductive and inductive settings.

With the transductive teacher, where seen and unseen classes are available, the generator can synthesise features of all classes. Given the test features, we can obtain the predicted class labels as follows:
\begin{equation}\label{10}
\setlength{\abovedisplayskip}{5pt} 
\setlength{\belowdisplayskip}{5pt}
y^*=\mathop{argmax}\limits_{y \in \mathcal Y}p(y|x,\theta_{S}^{*}), 
\end{equation}
where $\theta_{S}^{*}$ denotes optimised parameters of student. 

With the inductive teacher, where only seen class data is available, the problem is more challenging because the information of unseen classes is unavailable for both server and client during training. The generator is utilised to synthesise data of unseen classes. Given the noise $z$ and unseen class semantic embedding, the generated features can be obtained as $\tilde{x}=G(z|a; \theta_{G}^*)$. Then it is converted into a supervised learning task. The generated features are adopted to train a classifier $C$ as follows:
\begin{equation}\label{12}
\setlength{\abovedisplayskip}{5pt} 
\setlength{\belowdisplayskip}{5pt}
\min_{\theta_{C}}  -\mathbb{E}\left[ \log( y|\tilde{x}; \theta_{C})) \right].
\end{equation}
And class labels of test features can be predicted as follows:
\begin{equation}\label{13}
\setlength{\abovedisplayskip}{10pt} 
\setlength{\belowdisplayskip}{7pt}
y^*=\mathop{argmax}\limits_{y \in \tilde{\mathcal Y}}p(y|x,\theta_{C}^{*}),
\end{equation}
where $\mathcal{\tilde{Y}} = \mathcal Y_u$ for the conventional ZSL task, and $\mathcal{\tilde{Y}} = \mathcal Y_s \cup \mathcal Y_u$ for GZSL task.

\begin{table*}[t]
\caption{Detailed statistics of each dataset under data split in AZSL. Notation: `S' - seen class; `U' - unseen class; `Trans' - transductive teacher; `Ind' - inductive teacher. }
\begin{center}
\label{Dataset-ps}
\renewcommand\arraystretch{1}
\small
\resizebox{0.9\textwidth}{!}
{
%\addvbuffer[-5pt -28pt]{
%\setlength{\tabcolsep}{0.9mm}{
%\begin{tabular}{cccc|cc|cc|cc}
\begin{tabular}{m{1.5cm}<{\centering}m{2.2cm}<{\centering}m{2cm}<{\centering}m{1.2cm}<{\centering}|m{1.5cm}<{\centering}m{1.5cm}<{\centering}|m{1cm}<{\centering}m{1cm}<{\centering}|m{1cm}<{\centering}m{1.5cm}<{\centering}}
\toprule  %添加表格头部粗线

\textbf{Dataset}&   {\textbf{Semantics}} &\textbf{\textbf{Class Number}} &{\textbf{Image}}& \multicolumn{2}{c|}{\textbf{Teacher (Trans / Ind)}}& \multicolumn{2}{c|}{\textbf{AZSL Training}} & \multicolumn{2}{c}{\textbf{AZSL Evaluation}} \\
&&&& &&&&\multicolumn{2}{c}{\textbf{(Trans / Ind)}}\\

&&\textbf{S / U}& &\textbf{S} &\textbf{U} &   \textbf{S} &   \textbf{U} & \textbf{S} &\textbf{U}\\
\midrule  %添加表格中横线

AWA1 \cite{lampert2013attribute}  &BERT / attributes & 40 / 10 & 30475 & 19832&4542 / 0 & \textbf{0} &\textbf{0} & 4958 & 1143 / 5685\\
AWA2 \cite{xian2017zero}  &BERT / attributes  & 40 / 10 & 37322& 23527 & 6328 / 0 & \textbf{0} & \textbf{0} & 5882 & 1585 / 7913\\
aPY \cite{farhadi2009describing}  &BERT / attributes  & 20 / 12 & 15539 & 5932 & 6333 / 0 &  \textbf{0} & \textbf{0} & 1483 & 1591 / 7924\\

\bottomrule %添加表格底部粗线
\end{tabular}
}
%}
\end{center}
\end{table*}

\section{Experiments}

\paragraph{Datasets} 
We evaluate our AZSL model on three benchmark datasets: AWA1  \cite{lampert2013attribute}, AWA2 \cite{xian2017zero}) and aPY \cite{farhadi2009describing}. AWA1 and AWA2 consist of 30,475 and 37,322 images respectively and both datasets have 50 classes. aPY contains 15,539 images of 32 classes. In terms of semantic representation, we adopt the word embedding generated by language representation model BERT \cite{devlin2018bert} and the dimension of semantic embedding is 768 for all datasets. As for data split, there exist differences with two types of teacher models. In terms of inductive teacher, we follow the proposed data split proposed in \cite{xian2017zero}, \ie, only seen class data is available for the teacher. Transductive teacher is trained with all classes. We split the unseen classes randomly into training and test with a 4:1 ratio, which follows the proportion of seen data split in \cite{xian2017zero}. The detailed information of each dataset under data splits in AZSL is shown in Table \ref{Dataset-ps}.

\paragraph{Implementation Details}
As for image representation, We adopt the 2048-dimensional ResNet101 \cite{he2016deep} features following \cite{xian2017zero}. In our proposed framework, all networks are based on multi-Layer perceptrons (MLP) with the LeakyReLU activations \cite{Xu2015Empirical}. The teacher and student have the same architecture, which consists of two hidden layers with 1024 and 512 hidden units respectively. The generator contains a single hidden layer with 4096 hidden units and its output layer is ReLU. During the training process, we adopt the Adam \cite{kingma2014adam} optimiser and the learning rate of each network is set to $10^{-5}$. The dimension of the noise vector $z$ is hyper-parameter, which is empirically set to 20 for all datasets. The weight of regularization term is empirically set to 0.5 for AWA1 and AWA2, and 1 for aPY. The number of generated features is determined considering the trade-off between accuracy and computational efficiency. In practice, we generate 400 generated features in average per class for all datasets.

\paragraph{Evaluation Protocol}
We follow the evaluation metrics proposed in \cite{xian2017zero}. For conventional ZSL task, we use the per-class average top-1 accuracy to evaluate the classification performance to alleviate the data imbalance of classes. For GZSL task, we use harmonic mean $H=(2 \times u \times s)/(u+s)$ for evaluation, where $u$ and $s$ denote average per-class top-1 accuracy on unseen and seen classes, respectively. It is noteworthy that existing methods aim to classify unseen data into corresponding unseen classes in conventional ZSL task, while the class space at test time involves both unseen and seen classes in AZSL with transductive teacher. This makes AZSL with a transductive teacher more difficult compared with existing ZSL methods.

\begin{table*}[t]
    \caption{Comparison results with the state-of-the-art methods in both conventional ZSL and GZSL tasks. Conventional ZSL measures per-class average top-1 accuracy (T1) on unseen classes. GZSL measures u = T1 on unseen classes, s = T1 on seen classes, and H = harmonic mean. `WB’ represents white-box scenario. `BB’ represents black-box scenario. The upper part of the table contains inductive ZSL methods, the lower part contains transductive ZSL methods. The best results are in bold.}\label{AZSL-ALL}
\begin{center}    
\renewcommand\arraystretch{0.7}
\small
\resizebox{0.9\textwidth}{!}
{
%\addvbuffer[-3pt -10pt]{
\begin{tabular}{ccccc|ccc|ccc|ccc}
\toprule  %添加表格头部粗线
&Method & \multicolumn{3}{c|}{Zero-Shot Learning} & \multicolumn{9}{c}{Generalised Zero-Shot Learning}\\

 &&  AWA1& AWA2  & aPY  &\multicolumn{3}{c}{AWA1} &\multicolumn{3}{c}{AWA2} &\multicolumn{3}{c}{aPY}\\
&&T1&T1&T1& u & s &H &u & s &H & u & s &H \\

\midrule  %添加表格中横线
\multirow{11}{*}{Inductive}  
&IAP \cite{lampert2013attribute} &35.9&35.9&36.6 &2.1 &78.2 &4.1 &0.9 &87.6 &1.8 & 5.7 &65.6 &10.4\\ 
&DAP \cite{lampert2013attribute}  &  44.1& 46.1 & 33.8 &0.0 & \textbf{88.7} &0.0 &0.0 &84.7& 0.0  &4.8 &\textbf{78.3}&9.0\\
&ALE \cite{akata2013label}  &    59.9&    62.5    & 39.7 &16.8  &76.1  &27.5  &14.0  &81.8  &23.9  &4.6  &73.7  &8.7\\
&DEVISE \cite{frome2013devise}   &54.2&    59.7&   \textbf{39.8}&13.4     &68.7 &22.4 &17.1 &74.7 &27.8 &4.9 &76.9 &9.2\\
%CONSE \cite{norouzi2013zero}  &45.6    &44.5  &  26.9&0.4 &88.6 &0.8 &0.5 & \textbf{90.6} &1.0  &0.0 &\textbf{ 91.2} &0.0\\
%ESZSL \cite{romera2015embarrassingly}&58.2    &58.6&    38.3  &6.6 &75.6 &12.1 &5.9 &77.8 &11.0  &2.4& 70.1 &4.6\\

%SJE \cite{akata2015evaluation} &65.6&    61.9 &  32.9&11.3 &74.6 &19.6 &8.0 &73.9 &14.4  &3.7 &55.7 &6.9\\
%LATEM \cite{xian2016latent}  &55.1    &55.8 &   35.2  &7.3 &71.7 &13.3 &11.5 &77.3 &20.0  &0.1 &73.0 &0.2\\
&SYNC \cite{changpinyo2016synthesized} &54.0    &46.6&    23.9&8.9 &87.3 &16.2 &10.0 &\textbf{90.5} &18.0  &7.4 &66.3 &13.3\\
%SAE \cite{kodirov2017semantic}   &53.0&    54.1&   8.3 &1.8 &77.1 &3.5 &1.1 &82.2 &2.2  &0.4 &80.9 &0.9\\

&DEM \cite{zhang2017learning}  & 68.4 & 67.1& 35.0   & 32.8 & 84.7 & 47.3 & 30.5&  86.4 & 45.1  & 11.1 & 75.1 & 19.4 \\
%PSR \cite{annadani2018preserving} & - & 63.8 & 38.4  &-&-&- & 20.7 & 73.8 & 32.3  & 13.5 & 51.4 & 21.4 \\

&f-CLSWGAN \cite{xian2018feature}  & 68.2 & -  & -& 57.9 & 61.4 & 59.6 & - &-&- & -&-&-\\
&SE-GZSL \cite{verma2017generalized}  & 69.5 & 69.2  & - & 56.3 & 67.8 & 61.5& 58.3  & 68.1 & 62.8  & - & - & - \\
%cycle-CLSWGAN \cite{felix2018multi}   &66.8 &  - & - &56.9 &64.0 &60.2 &-&-&-& -&-&- \\
%DLFZRL \cite{tong2019hierarchical}&  66.3 & 63.7 &44.5 & -&-& 40.5&-&- &45.1&-&- &31.0 \\
%GFZSL \cite{verma2017simple}  & 68.3 &63.8 &  38.4 &  1.8& 80.3 &3.5 & 2.5 &80.1 &4.8 & 0.0 &83.3 &0.0 \\
%GDAN \cite{huang2019generative}  &-&-&-&-&-&-& 33.2 & 67.5 & 44.6 & 30.4& 75.0& 43.4  \\
%CVAE-ZSL \cite{mishra2018generative} &  71.4 & 65.8 &- & -&-&47.2 &-&-&51.2 &-&-&- \\
%DE-VAE \cite{ma2020variational}& 69.4 & 69.3 &- & 59.6 &76.1& 66.9 & 58.8& 78.9& 67.4&-&-&-\\
&ISE-GAN \cite{pambala2020generative} & 68.4 &65.6 &- & 58.7 &74.4 & 65.6 &55.9&79.3  &65.5&-&-&-\\
&Disentangled-VAE \cite{li2021generalized} &-&-&- &60.7& 72.9 &66.2& 56.9& 80.2& 66.6&-&-&-\\
&CE-GZSL \cite{han2021contrastive} & \textbf{71.0} &\textbf{70.4}&-& \textbf{65.3} &73.4 &\textbf{69.1 }&\textbf{63.1 }&78.6 &\textbf{70.0}&-&-&-\\
%\vspace{1ex}
%\cline{2-14}
\midrule 
&\textbf{AZSL+BB} &14.1 & 19.9 & 12.3 &   4.1 & 3.7& 3.9 & 3.5&3.7 &3.6& 6.8 &4.0 & 5.1\\
%\textbf{White-Box}&34.1 & 33.0&12.8 & 16.4&47.8&24.5&22.6&63.0&33.3 & 15.4&54.2&24.0 \\
&\textbf{AZSL+WB} & 34.5 & 36.5 & 18.7 & 23.4 &34.3 &27.8 & 27.3 &44.3 &33.7 & \textbf{17.9 }& 52.5 &\textbf{26.7} \\
%\vspace{1ex}
\midrule 

\multirow{7}{*}{Transductive}
&QFSL \cite{song2018transductive} & -&79.1&- &-&-&-&66.2&93.1&77.4 &-&-&- \\
&PREN \cite{ye2019progressive}  &-&78.6&-&-&-&-& 32.4& 88.6& 47.4 &-&-&-\\
&GMN \cite{sariyildiz2019gradient} &82.5 &-&-&70.8& 79.2 &74.8&-&-&-\\
&DTN \cite{zhang2020deep} & 69.0 &-& 41.5 & 54.8 & 88.5 &67.7 &-&-&- &37.4 &\textbf{87.9} &52.5\\ 
&GMSADE \cite{gune2020generative} &81.3 &\textbf{80.7 }&49.9 & 71.2  &87.7& 78.6 &71.3&86.1&  78.0 &76.1 &39.3 &51.8\\
&EDE \cite{zhang2020towards} & \textbf{85.3} &77.5& 31.3 & 71.4 &\textbf{90.1}& 79.7 &68.4 &\textbf{93.2 }&78.9&29.8 &79.4 &43.3\\
\midrule 
%\vspace{1ex}
%\cline{2-14}
%\textbf{Black Box}& 25.6 &15.6& 19.4&24.7&18.1&20.9&11.8&20.9&15.0\\
%\textbf{White-Box}&\textbf{75.8}& \textbf{76.1} & \textbf{83.0}&\textbf{ 75.8}& 82.3&\textbf{78.9} &\textbf{76.1}&83.8&\textbf{79.8} &\textbf{83.0}&84.5&\textbf{83.7}\\
&\textbf{AZSL+BB} & 33.5 & 29.0 &30.2 &33.5 & 28.6 & 30.9 &29.0 & 25.3& 27.0& 30.2&42.2 & 35.2 \\
&\textbf{AZSL+WB} &77.9 & 79.0& \textbf{83.9} & \textbf{77.9} &81.8&\textbf{79.8 }& \textbf{79.0}&86.7&\textbf{82.7} & \textbf{83.9}& 85.7 &\textbf{84.8} \\

\bottomrule %添加表格底部粗线
\end{tabular}
}
%}
\end{center}
\end{table*}

\begin{table*}[!t]
\begin{center}    
\caption{Experimental results in black-box scenario with transductive teacher in both conventional ZSL and GZSL task. `BB’ represents black-box scenario. $u$ = Top-1 accuracy (T1) on unseen classes, $s$ = Top-1 accuracy (T1) on seen classes, and $H$ = harmonic mean.}
\label{AZSL-all-blackbox}
\renewcommand\arraystretch{0.7}
\small
\resizebox{0.8\textwidth}{!}
{
%\addvbuffer[5pt -18pt]{
\begin{tabular}{cccc|ccc|ccc|ccc}
\toprule  %添加表格头部粗线
Method & \multicolumn{3}{c|}{Zero-Shot Learning} & \multicolumn{9}{c}{Generalised Zero-Shot Learning}\\

 &  AWA1& AWA2  & aPY  &\multicolumn{3}{c}{AWA1} &\multicolumn{3}{c}{AWA2} &\multicolumn{3}{c}{aPY}\\
&T1&T1&T1& u & s &H &u & s &H & u & s &H \\

\midrule  %添加表格中横线
%\textbf{Pure Noise}  \\
Label-Conditioned & 15.5&10.0 & 7.0&15.5&24.3& 18.9&10.0&17.8&12.8&7.0& 3.8&4.9\\
Attribute-Conditioned & 10.1 &23.0& 8.2 &10.1& 11.3&10.7&23.0&17.6&20.0&8.2& 5.0 & 6.3\\
w/o Data Verification & 25.6 & 24.7 & 11.8 & 25.6 &15.6& 19.4&24.7&18.1&20.9&11.8&20.9&15.0\\
w/o Regularization &26.8 & 23.7&23.2 & 26.8 &26.7&26.8&23.7&23.2 & 23.4& 23.2&25.6&24.3\\
\midrule  %添加表格中横线
\textbf{AZSL+BB} & \textbf{33.5} & \textbf{29.0 }&\textbf{30.2 }&\textbf{ 33.5} & \textbf{28.6} & \textbf{30.9 }&\textbf{29.0} & \textbf{25.3 }&\textbf{27.0 }& \textbf{30.2}&\textbf{ 42.2} & \textbf{35.2} \\

\bottomrule %添加表格底部粗线
\end{tabular}
%}
}
\end{center}
\end{table*}

\subsection{Main Results}

\paragraph{Comparisons with State-of-Art Methods}

We present experimental results of the proposed AZSL model in both conventional and generalised ZSL task in Table \ref{AZSL-ALL}. Considering this is the first proposed AZSL work, we provide comparison with traditional state-of-the-art ZSL methods as a reference. Note that all of our AZSL models are at significant disadvantage since no real data is involved during training. Instead, the AZSL model generates 400 features using class semantic embedding only for each class and request guidance feedback from the teacher model.

%AZSL model is guided by the teacher via weight sharing and only output guidance in white-box and black-box scenario respectively. In terms of different types of teachers, teacher can access data of all classes and inductive teacher only obtains seen class information.%, \ie, all the framework cannot access any information of unseen classes during training. 

Our initial assumption was that if the generated feature can approximate the real data distribution, the performance should be close to those methods using real data. To our surprise, AZSL model with transductive teacher achieves promising performance in both conventional and generalised ZSL in white-box scenario despite the disadvantage in no access to the real data. Our model achieves the best performance in GZSL on all three datasets especially on aPY, with 32.3\% increases in harmonic mean, indicating the better balance on seen and unseen classes of our AZSL model. In conventional ZSL, AZSL model gains the best performance on aPY and has competitive classification accuracy on AWA1 and AWA2. It is noteworthy that the class space contains all classes during the testing process, so it is more challenging in conventional ZSL task. As for black-box scenario, AZSL model shows good potential to achieve good balance between seen and unseen classes. For example, the accuracy on unseen classes is higher than seen classes on AWA2, with 4.9\% higher performance. It indicates that AZSL model is promising to mitigate class-level overfitting issue \cite{zhang2018triple} in GZSL task.

Compared with inductive ZSL methods, results show that our proposed model with inductive teacher in white-box scenario gains the satisfactory performance in GZSL especially on aPY, with 7.3\% higher performance on harmonic mean of our AZSL model. Although the improvement on inductive setting is not significant, it is very impressive for us that the student network can generate new knowledge beyond the source data of teacher model. For black-box scenario, the results show that our AZSL model outperforms random guessing, which are around 10\% on AWA1, AWA2 and 8\% on aPY according to unseen class number. The white-box achieves better performance than black-box, indicating that the gradient guidance provides more information.

\iffalse
\begin{table*}[!t]
\begin{center}    
\caption{Experimental results with inductive teacher in both CZSL and GZSL task. `WB’ and `BB' represents white-box and black- box scenarios respectively. $u$ = Top-1 accuracy (T1) on unseen classes, $s$ = Top-1 accuracy (T1) on seen classes, and $H$ = harmonic mean.}
\label{AZSL}
\renewcommand\arraystretch{0.7}
\footnotesize
\resizebox{0.8\textwidth}{!}
{
%\addvbuffer[2pt -10pt]{
\begin{tabular}{cccc|ccc|ccc|ccc}
\toprule  %添加表格头部粗线
Method & \multicolumn{3}{c|}{Zero-Shot Learning} & \multicolumn{9}{c}{Generalised Zero-Shot Learning}\\

 &  AWA1& AWA2  & aPY &\multicolumn{3}{c}{AWA1} &\multicolumn{3}{c}{AWA2} &\multicolumn{3}{c}{aPY}\\
&T1&T1&T1& u & s &H &u & s &H & u & s &H \\

\midrule  %添加表格中横线

%\textbf{Black-Box}& 10.4 &11.4 & 9.9 & 2.7&1.7&2.1 & 5.1&2.4&3.3 &2.0&4.6&2.8\\
\textbf{AZSL+BB} &14.1 & 19.9 & 12.3 &   4.1 & 3.7& 3.9 & 3.5&3.7 &3.6& 6.8 &4.0 & 5.1\\
%\textbf{White-Box}&34.1 & 33.0&12.8 & 16.4&47.8&24.5&22.6&63.0&33.3 & 15.4&54.2&24.0 \\
\textbf{AZSL+WB} & 34.5 & 36.5 & 18.7 & 23.4 &34.3 &27.8 & 27.3 &44.3 &33.7 & 17.9 & 52.5 &26.7 \\

\bottomrule %添加表格底部粗线
\end{tabular}
%}
}
\end{center}
\end{table*}

\fi

\noindent\textbf{Comparisons in Black-Box Scenario}
In this situation, transductive teacher contains the information of all classes and it gives the output guidance during training. As it is the first time to propose this setting, we provide several baselines for comparison shown in Table \ref{AZSL-all-blackbox}. We provide several types of class embedding for conditional feature generation. `Label-' and `attribute-conditioned' represent the feature synthesis conditioned on class label and attributes respectively. Compared with these two baselines, our proposed framework with BERT embedding achieves the best performance, \ie, with 18.0\% and 23.4\% increases in harmonic mean on AWA1. Besides, results show that our framework gains obvious improvement in accuracy with data verification, \ie, with 20.2\% higher performance on harmonic mean on aPY dataset. In terms of feature generation constraint, the results indicate the effectiveness to adopt regularization, \ie, it achieves 4.4\% and 10.9\% increases in Harmonic mean on AWA2 and aPY. The comparison with baseline models demonstrate the effectiveness of our AZSL model in black-box scenario with transductive teacher. 

\begin{table}[!t]
\begin{center}    
\caption{Experimental results with different constraints for feature generation in GZSL task. ‘CE’ represents cross-entropy loss, `MMD' represents MMD distance loss, `KL' represents KL divergence loss. $u$ = Top-1 accuracy on unseen classes, $s$ = Top-1 accuracy on seen classes, and $H$ = harmonic mean.}
\label{AZSL-loss}
\renewcommand\arraystretch{0.7}
\small
\resizebox{0.45\textwidth}{!}
{
%\addvbuffer[0pt -5pt]{
\begin{tabular}{cccc|ccc}
\toprule  %添加表格头部粗线
Method 
 &\multicolumn{3}{c}{AWA2} &\multicolumn{3}{c}{aPY}\\
  &u & s &H  & u & s &H \\

\midrule  %添加表格中横线

CE&76.1&83.8&79.8 &83.0&84.5&83.7\\
CE+MMD& \textbf{79.9} & 85.1 & 82.5 & 81.5 & 85.5 & 83.5  \\
CE+KL & 79.0&\textbf{86.7}&\textbf{82.7} & \textbf{83.9}& \textbf{85.7 }&\textbf{84.8}\\

\bottomrule %添加表格底部粗线
\end{tabular}
%}
}
\end{center}
\end{table}

\subsection{Discussion and Limitations}

\paragraph{Feature Generation Regularization Analysis}
The key issue in our data-free knowledge transfer framework is to generate high-quality features, which are expected to have similar distribution to real data. To show the influence of different constraints during feature generation process, we provide analysis with different regularization term for generator training in Table \ref{AZSL-loss}. KL and MMD loss \cite{gretton2012kernel} aim to minimise the distribution difference between real and generated features. The results show that adding the distribution constraint of synthesised data is beneficial for feature generation. For example, the harmonic mean increases 2.7\% and 2.9\% with MMD and KL loss respectively compared with the baseline that only contains cross entropy loss. Besides, the results indicate that KL loss and MMD loss are both effective in improving the quality of generated data and KL loss performs better to a small extent, which shows the effectiveness of KL regularization.

\begin{figure}[t]
    \centering
    \subfloat[AWA1]{
        \includegraphics[width=0.22\textwidth]{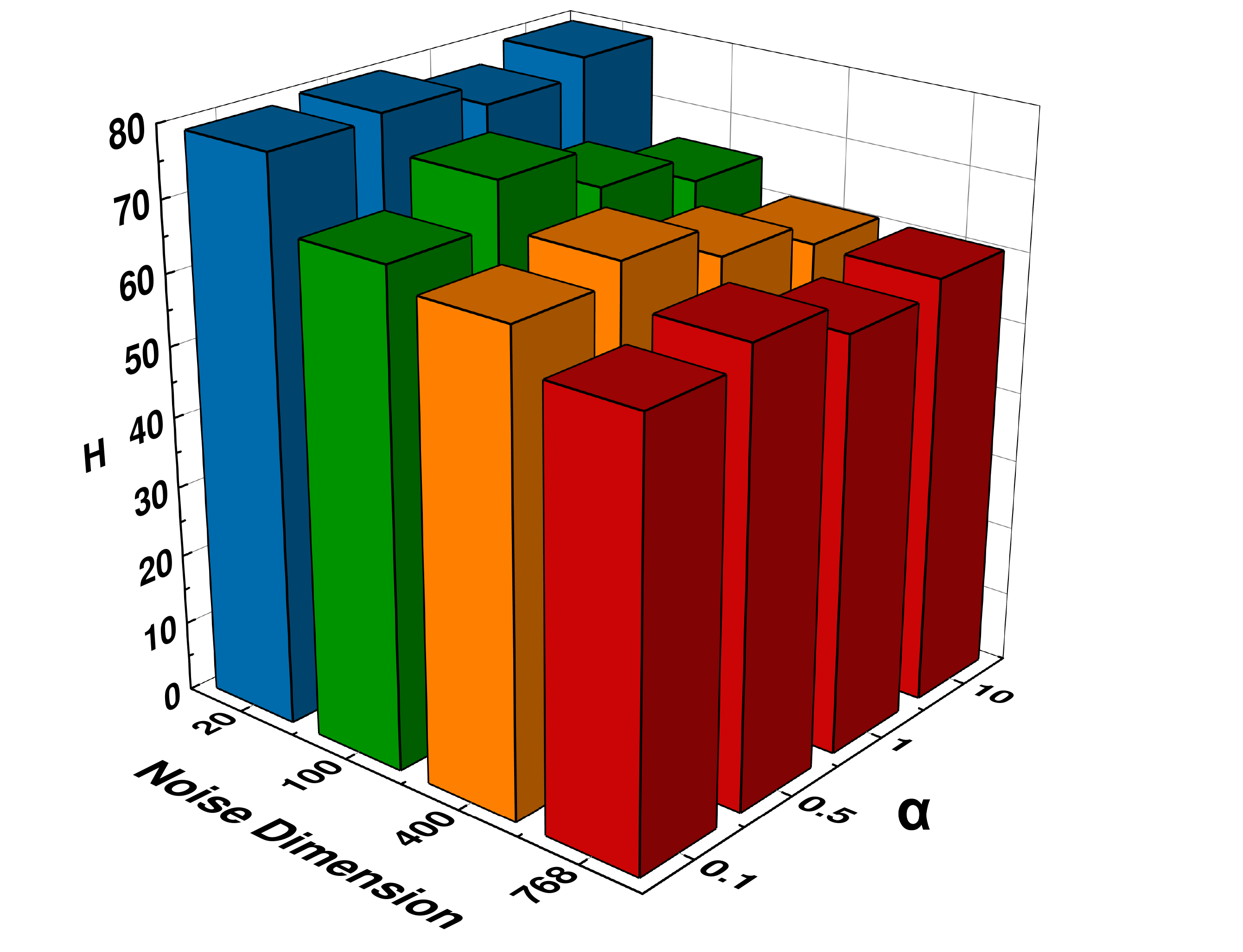}
     }
    \subfloat[aPY]{
        \includegraphics[width=0.22\textwidth]{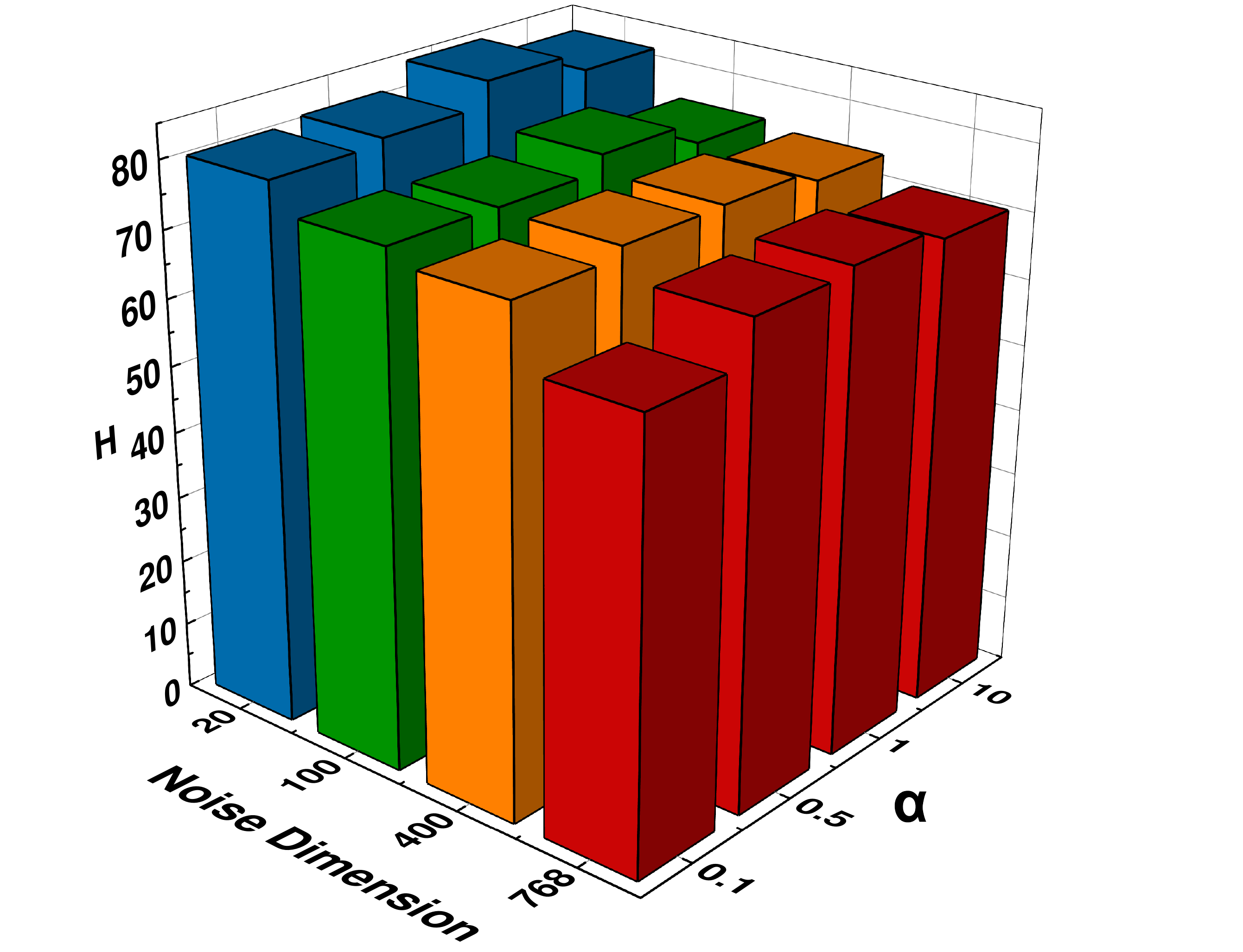}
    }
    \caption{Harmonic mean in GZSL with different noise dimension and parameter $\alpha$ on AWA1 and aPY with transductive teacher in white-box scenario.}\label{hyper} 
\end{figure}

\begin{figure}[t]
    \centering
    \subfloat[AWA1]{
        \includegraphics[width=0.22\textwidth]{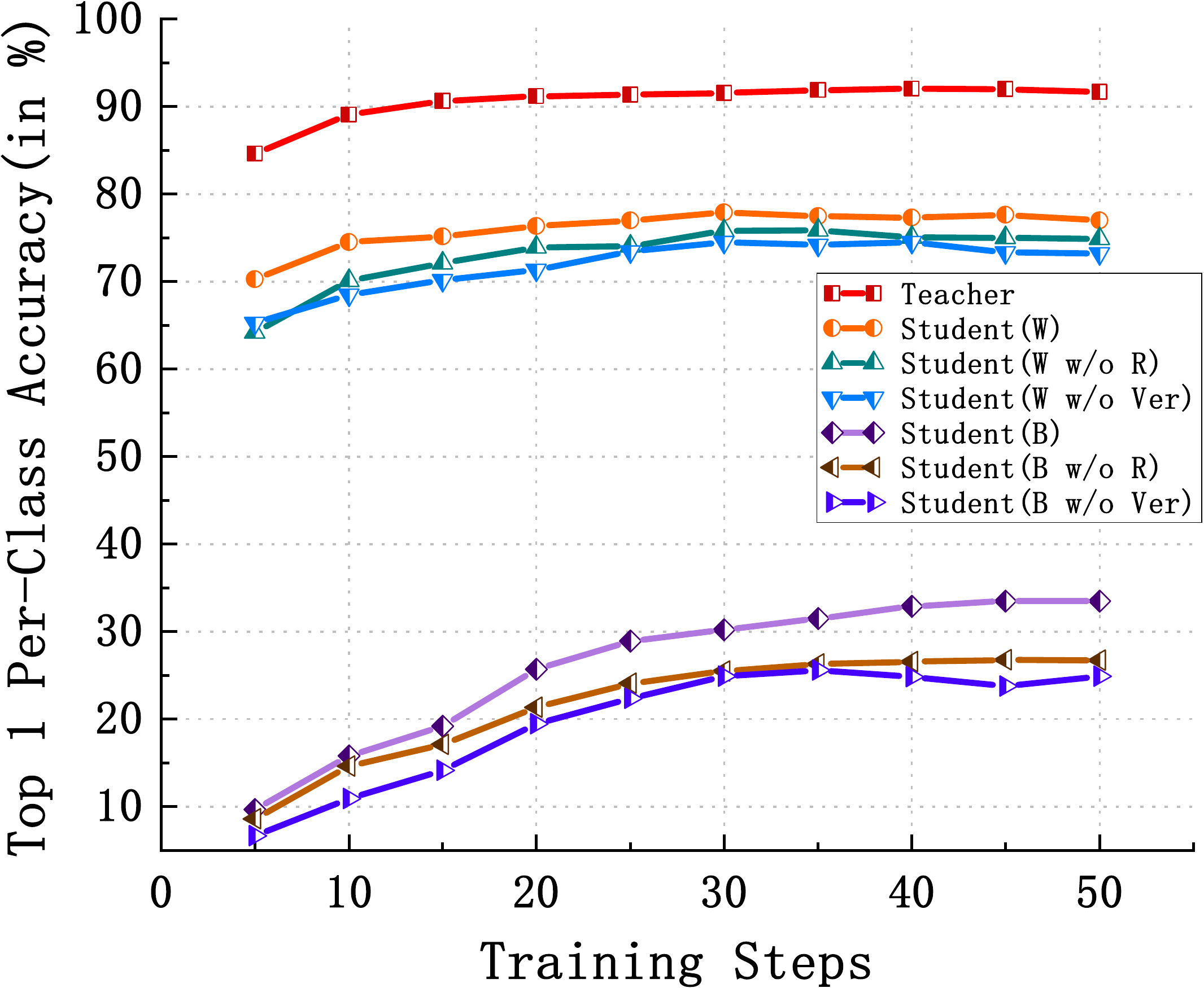}
     }
    \subfloat[aPY]{
        \includegraphics[width=0.22\textwidth]{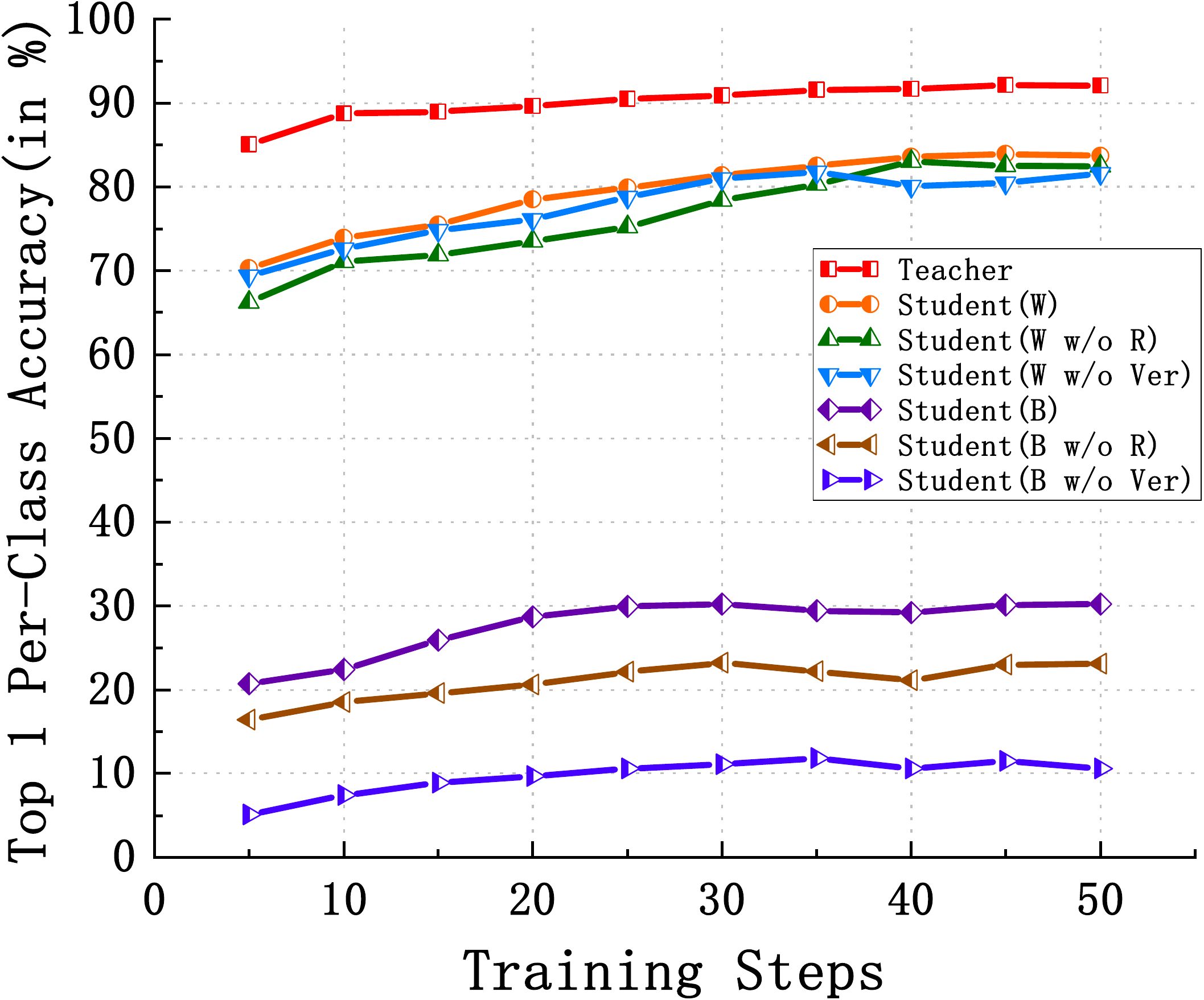}
    }
    \caption{Unseen class accuracy with increasing epochs of student with transductive teacher on AWA1 and aPY. `W' and `B' represent white-box and black-box respectively. `Ver' represents data verification process. `R' represents regularization term.}
    \label{acc-t-s}
\end{figure}

\noindent\textbf{Hyper-Parameter Analysis} We evaluate the influence of two hyper-parameters, \ie noise dimension and regularization weight, in our model. We have two ablation studies on AWA1 and aPY based white-box scenario with transductive teacher in Figure \ref{hyper}. We choose four different noise dimensions, \ie, 20,100,400,768, to show the relationship with harmonic mean. Results show that the performance decreases with the increase of noise dimension on both datasets, indicating that noise of high dimension may bring much interference. In terms of regularization weight, We set $\alpha$ = 0.01, 0.1, 1, 10 in the experiments. As shown in Figure \ref{hyper}, harmonic mean on both datasets change slightly with different $\alpha$. And the best performance can be achieved when $\alpha$ is 0.5 and 1 on AWA1 and aPY respectively.

\noindent\textbf{Student Performance Analysis} Here we conduct some experiments to analyze the student performance. We present classification accuracy of teacher and student networks with increasing training steps in both scenarios with transductive teacher on AWA1 and aPY. The teacher acts as the upper bound so student aims to achieve similar performance trained on generated features. As shown in Figure \ref{acc-t-s}, student in white-box scenario obtains the satisfactory result close to teacher, indicating that gradient guidance is effective for framework training. Besides, results show that model achieves better performance with regularization term, indicating the effectiveness of feature distribution during training. And statistics also show that framework performs better with data verification in both scenarios, which indicates its necessity because it can mitigate the negative influence caused by generated features of inferior quality.

\begin{figure}
    \centering
    \subfloat[AWA1 Real Features]{
        \includegraphics[width=0.22\textwidth]{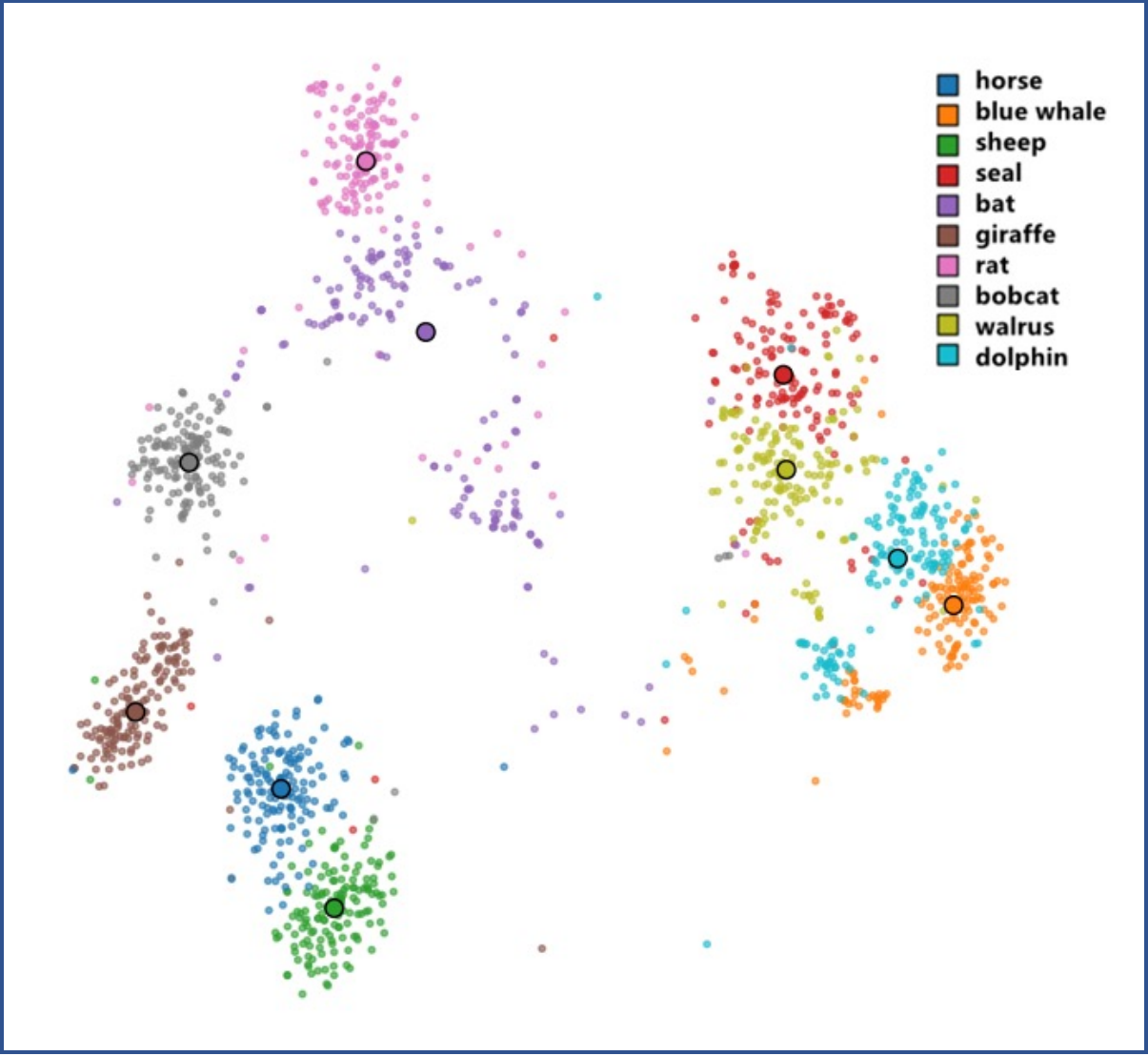}
     }
    \subfloat[AWA1 Generated Features]{
        \includegraphics[width=0.22\textwidth]{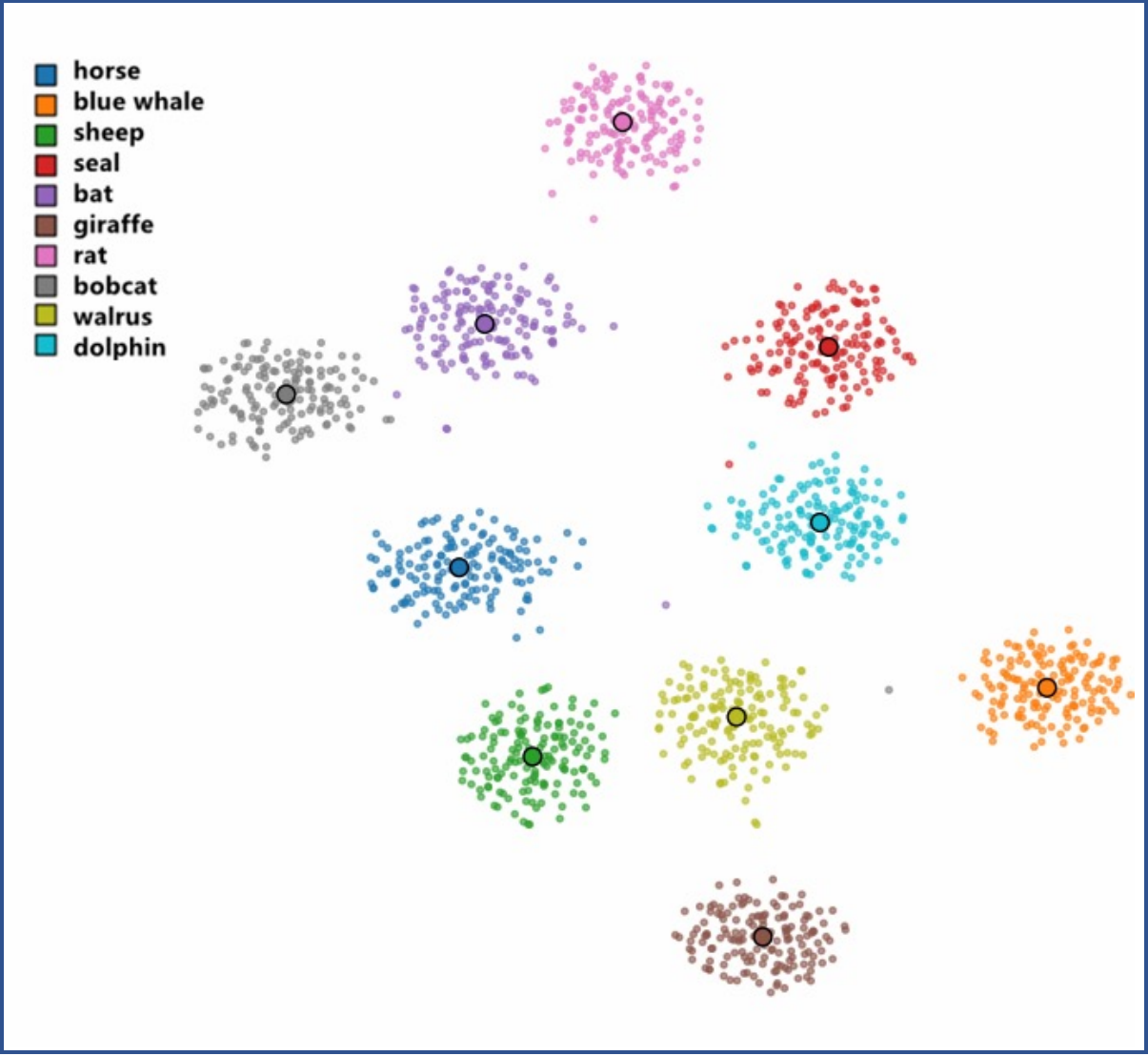}
    }
    
    \quad
        \subfloat[aPY Real Features]{
        \includegraphics[width=0.22\textwidth]{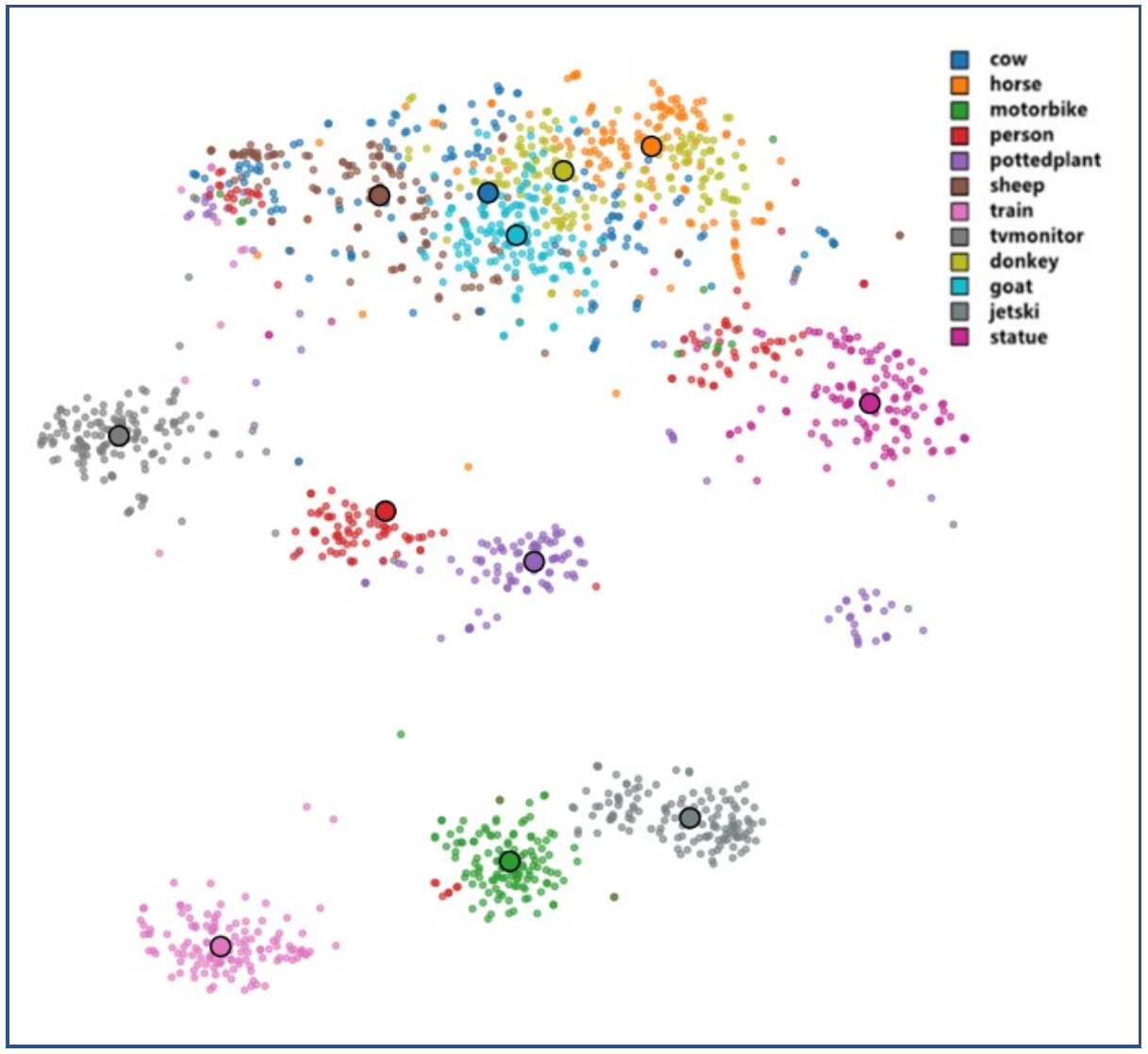}
     }
    \subfloat[aPY Generated Features]{
        \includegraphics[width=0.22\textwidth]{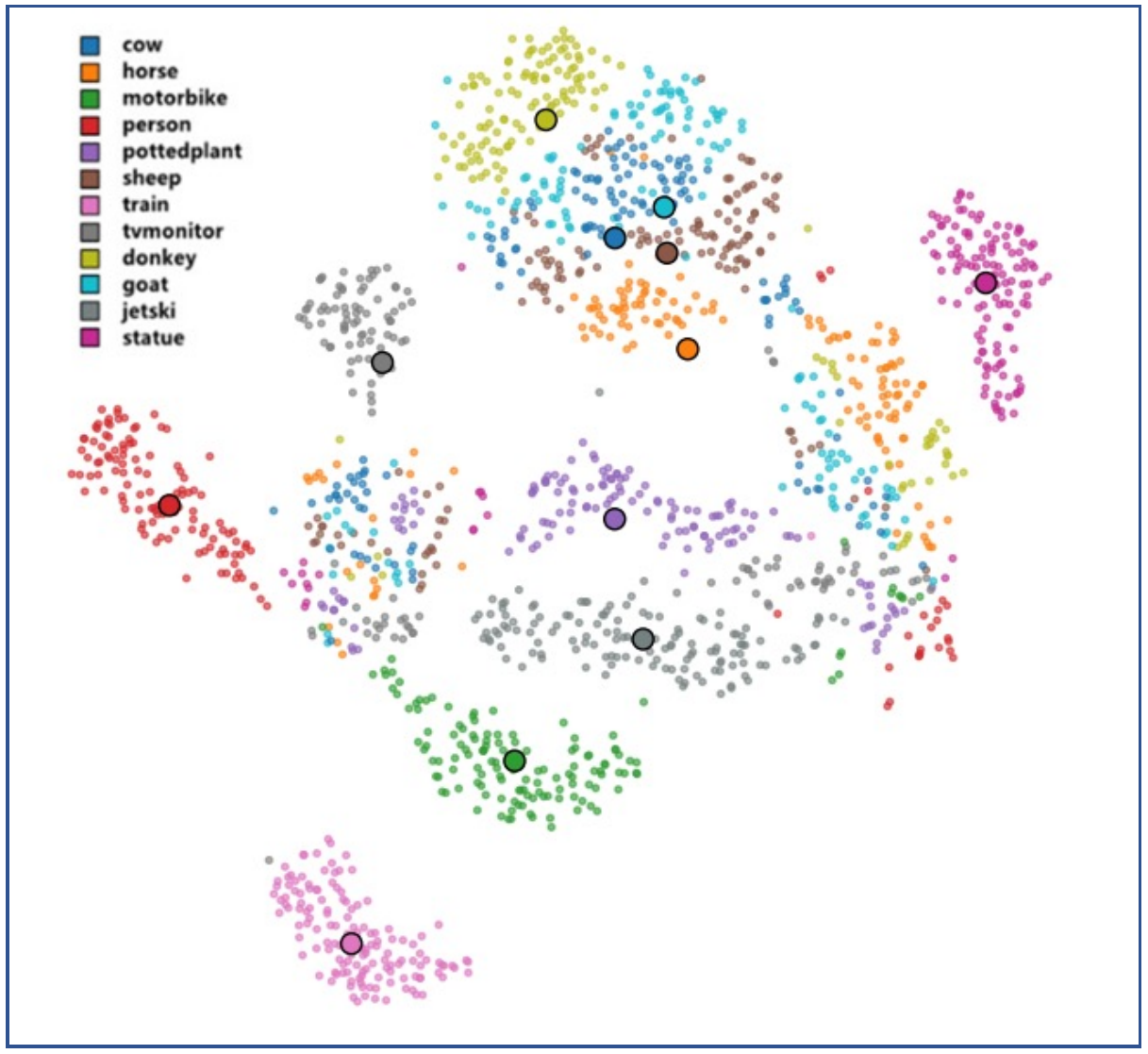}
    }
    
    \caption{t-SNE visualization of the real and synthesised features of unseen classes in white-box scenario with transductive teacher on AWA1 and aPY dataset. }
    \label{TSNE}
\end{figure}

\noindent\textbf{Quality of Generated Features} Figure \ref{TSNE} shows the t-SNE visualization of real and generated unseen class features in both scenarios with transductive teacher on AWA1 and aPY dataset. We randomly choose a part of features for clear visualization. The generated features have distribution close to the real ones. And generated features are more class-level clustered than real features, indicating the effectiveness of feature generation under the supervision of teacher guidance, even though real data is unavailable. Therefore, generated features can be viewed as a suitable replacement for the real features.

\section{Conclusions}
\noindent This paper has presented a privacy preserving AZSL paradigm. A pre-trained teacher model was deployed on the data owner server as the data safeguard to provide guidance for AZSL training, in terms of data generation and verification. We extensively studied the proposed `black-box' and `white-box' scenarios and their trade-off in performance and security. Despite an embarrassingly simple MLP network, our framework maintains promising performance in both CZSL and GZSL tasks despite the absence of real data during training. The AZSL paradigm protected the data privacy and meanwhile mitigated the seen-unseen bias in GZSL tasks. Future development of AZSL can focus on model design, research communication cost, security of black-box and white-box scenario, and improving the performance on inductive AZSL settings.

%%%%%%%%% REFERENCES
\bibliography{azsl}

\begin{thebibliography}{57}
\providecommand{\natexlab}[1]{#1}

\bibitem[{Akata et~al.(2013)Akata, Perronnin, Harchaoui, and
  Schmid}]{akata2013label}
Akata, Z.; Perronnin, F.; Harchaoui, Z.; and Schmid, C. 2013.
\newblock Label-embedding for attribute-based classification.
\newblock In \emph{CVPR}.

\bibitem[{Akata et~al.(2015)Akata, Reed, Walter, Lee, and
  Schiele}]{akata2015evaluation}
Akata, Z.; Reed, S.; Walter, D.; Lee, H.; and Schiele, B. 2015.
\newblock Evaluation of output embeddings for fine-grained image
  classification.
\newblock In \emph{CVPR}.

\bibitem[{Annadani and Biswas(2018)}]{annadani2018preserving}
Annadani, Y.; and Biswas, S. 2018.
\newblock Preserving semantic relations for zero-shot learning.
\newblock In \emph{CVPR}.

\bibitem[{Changpinyo et~al.(2016)Changpinyo, Chao, Gong, and
  Sha}]{changpinyo2016synthesized}
Changpinyo, S.; Chao, W.-L.; Gong, B.; and Sha, F. 2016.
\newblock Synthesized classifiers for zero-shot learning.
\newblock In \emph{CVPR}.

\bibitem[{Chao et~al.(2016)Chao, Changpinyo, Gong, and Sha}]{chao2016empirical}
Chao, W.-L.; Changpinyo, S.; Gong, B.; and Sha, F. 2016.
\newblock An empirical study and analysis of generalized zero-shot learning for
  object recognition in the wild.
\newblock In \emph{ECCV}.

\bibitem[{Chen et~al.(2021)Chen, Wang, Gan, Liu, Henao, and
  Carin}]{chen2021wasserstein}
Chen, L.; Wang, D.; Gan, Z.; Liu, J.; Henao, R.; and Carin, L. 2021.
\newblock Wasserstein contrastive representation distillation.
\newblock In \emph{Proceedings of the IEEE/CVF Conference on Computer Vision
  and Pattern Recognition}, 16296--16305.

\bibitem[{Devlin et~al.(2018)Devlin, Chang, Lee, and
  Toutanova}]{devlin2018bert}
Devlin, J.; Chang, M.-W.; Lee, K.; and Toutanova, K. 2018.
\newblock Bert: Pre-training of deep bidirectional transformers for language
  understanding.
\newblock \emph{arXiv preprint arXiv:1810.04805}.

\bibitem[{Farhadi et~al.(2009)Farhadi, Endres, Hoiem, and
  Forsyth}]{farhadi2009describing}
Farhadi, A.; Endres, I.; Hoiem, D.; and Forsyth, D. 2009.
\newblock Describing objects by their attributes.
\newblock In \emph{CVPR}.

\bibitem[{Felix et~al.(2018)Felix, Kumar, Reid, and Carneiro}]{felix2018multi}
Felix, R.; Kumar, V.~B.; Reid, I.; and Carneiro, G. 2018.
\newblock Multi-modal cycle-consistent generalized zero-shot learning.
\newblock In \emph{ECCV}.

\bibitem[{Frome et~al.(2013)Frome, Corrado, Shlens, Bengio, Dean, Mikolov
  et~al.}]{frome2013devise}
Frome, A.; Corrado, G.~S.; Shlens, J.; Bengio, S.; Dean, J.; Mikolov, T.;
  et~al. 2013.
\newblock Devise: A deep visual-semantic embedding model.
\newblock In \emph{NeurIPS}.

\bibitem[{Fu et~al.(2015)Fu, Hospedales, Xiang, and Gong}]{fu2015transductive}
Fu, Y.; Hospedales, T.~M.; Xiang, T.; and Gong, S. 2015.
\newblock Transductive multi-view zero-shot learning.
\newblock \emph{IEEE transactions on pattern analysis and machine
  intelligence}, 37(11): 2332--2345.

\bibitem[{Gao et~al.(2020)Gao, Hou, Qin, Chen, Liu, Zhu, Zhang, and
  Shao}]{gao2020zero}
Gao, R.; Hou, X.; Qin, J.; Chen, J.; Liu, L.; Zhu, F.; Zhang, Z.; and Shao, L.
  2020.
\newblock Zero-VAE-GAN: Generating Unseen Features for Generalized and
  Transductive Zero-Shot Learning.
\newblock \emph{IEEE Transactions on Image Processing}, 29: 3665--3680.

\bibitem[{Gretton et~al.(2012)Gretton, Borgwardt, Rasch, Sch{\"o}lkopf, and
  Smola}]{gretton2012kernel}
Gretton, A.; Borgwardt, K.~M.; Rasch, M.~J.; Sch{\"o}lkopf, B.; and Smola, A.
  2012.
\newblock A kernel two-sample test.
\newblock \emph{The Journal of Machine Learning Research}, 13(1): 723--773.

\bibitem[{Gune et~al.(2020)Gune, Pal, Mukherjee, Banerjee, and
  Chaudhuri}]{gune2020generative}
Gune, O.; Pal, M.; Mukherjee, P.; Banerjee, B.; and Chaudhuri, S. 2020.
\newblock Generative model-driven structure aligning discriminative embeddings
  for transductive zero-shot learning.
\newblock \emph{arXiv preprint arXiv:2005.04492}.

\bibitem[{Han et~al.(2021)Han, Fu, Chen, and Yang}]{han2021contrastive}
Han, Z.; Fu, Z.; Chen, S.; and Yang, J. 2021.
\newblock Contrastive Embedding for Generalized Zero-Shot Learning.
\newblock In \emph{CVPR}.

\bibitem[{Hao et~al.(2021)Hao, El-Khamy, Lee, Zhang, Liang, Chen, and
  Duke}]{hao2021towards}
Hao, W.; El-Khamy, M.; Lee, J.; Zhang, J.; Liang, K.~J.; Chen, C.; and Duke,
  L.~C. 2021.
\newblock Towards Fair Federated Learning with Zero-Shot Data Augmentation.
\newblock In \emph{CVPR}.

\bibitem[{He et~al.(2016)He, Zhang, Ren, and Sun}]{he2016deep}
He, K.; Zhang, X.; Ren, S.; and Sun, J. 2016.
\newblock Deep residual learning for image recognition.
\newblock In \emph{CVPR}.

\bibitem[{Jayaraman and Grauman(2014)}]{jayaraman2014zero}
Jayaraman, D.; and Grauman, K. 2014.
\newblock Zero-shot recognition with unreliable attributes.
\newblock In \emph{NeurIPS}.

\bibitem[{Kingma and Ba(2015)}]{kingma2014adam}
Kingma, D.~P.; and Ba, J. 2015.
\newblock Adam: A method for stochastic optimization.
\newblock In \emph{ICLR}.

\bibitem[{Kodirov et~al.(2015)Kodirov, Xiang, Fu, and
  Gong}]{kodirov2015unsupervised}
Kodirov, E.; Xiang, T.; Fu, Z.; and Gong, S. 2015.
\newblock Unsupervised domain adaptation for zero-shot learning.
\newblock In \emph{ICCV}, 2452--2460.

\bibitem[{Kodirov, Xiang, and Gong(2017)}]{kodirov2017semantic}
Kodirov, E.; Xiang, T.; and Gong, S. 2017.
\newblock Semantic Autoencoder for Zero-Shot Learning.
\newblock In \emph{CVPR}.

\bibitem[{Kone{\v{c}}n{\`y} et~al.(2016)Kone{\v{c}}n{\`y}, McMahan, Yu,
  Richt{\'a}rik, Suresh, and Bacon}]{konevcny2016federated}
Kone{\v{c}}n{\`y}, J.; McMahan, H.~B.; Yu, F.~X.; Richt{\'a}rik, P.; Suresh,
  A.~T.; and Bacon, D. 2016.
\newblock Federated learning: Strategies for improving communication
  efficiency.
\newblock \emph{arXiv preprint arXiv:1610.05492}.

\bibitem[{Lampert, Nickisch, and Harmeling(2009)}]{lampert2009learning}
Lampert, C.~H.; Nickisch, H.; and Harmeling, S. 2009.
\newblock Learning to detect unseen object classes by between-class attribute
  transfer.
\newblock In \emph{CVPR}.

\bibitem[{Lampert, Nickisch, and Harmeling(2013)}]{lampert2013attribute}
Lampert, C.~H.; Nickisch, H.; and Harmeling, S. 2013.
\newblock Attribute-based classification for zero-shot visual object
  categorization.
\newblock \emph{IEEE transactions on pattern analysis and machine
  intelligence}, 36(3): 453--465.

\bibitem[{Larochelle, Erhan, and Bengio(2008)}]{larochelle2008zero}
Larochelle, H.; Erhan, D.; and Bengio, Y. 2008.
\newblock Zero-data learning of new tasks.
\newblock In \emph{AAAI}.

\bibitem[{Li et~al.(2021)Li, Xu, Wei, and Deng}]{li2021generalized}
Li, X.; Xu, Z.; Wei, K.; and Deng, C. 2021.
\newblock Generalized Zero-Shot Learning via Disentangled Representation.
\newblock In \emph{AAAI}.

\bibitem[{Liu et~al.(2017)Liu, Dai, Humayun, Tay, Yu, Smith, Rehg, and
  Song}]{liu2017iterative}
Liu, W.; Dai, B.; Humayun, A.; Tay, C.; Yu, C.; Smith, L.~B.; Rehg, J.~M.; and
  Song, L. 2017.
\newblock Iterative machine teaching.
\newblock In \emph{International Conference on Machine Learning}, 2149--2158.
  PMLR.

\bibitem[{Long et~al.(2017)Long, Liu, Shao, Shen, Ding, and Han}]{long2017zero}
Long, Y.; Liu, L.; Shao, L.; Shen, F.; Ding, G.; and Han, J. 2017.
\newblock From zero-shot learning to conventional supervised classification:
  Unseen visual data synthesis.
\newblock In \emph{CVPR}.

\bibitem[{Long and Shao(2017)}]{long2017describing}
Long, Y.; and Shao, L. 2017.
\newblock Describing unseen classes by exemplars: Zero-shot learning using
  grouped simile ensemble.
\newblock In \emph{WACV}, 907--915. IEEE.

\bibitem[{Ma and Hu(2020)}]{ma2020variational}
Ma, P.; and Hu, X. 2020.
\newblock A Variational Autoencoder with Deep Embedding Model for Generalized
  Zero-Shot Learning.
\newblock In \emph{AAAI}.

\bibitem[{McMahan et~al.(2017)McMahan, Moore, Ramage, Hampson, and
  y~Arcas}]{mcmahan2017communication}
McMahan, B.; Moore, E.; Ramage, D.; Hampson, S.; and y~Arcas, B.~A. 2017.
\newblock Communication-efficient learning of deep networks from decentralized
  data.
\newblock In \emph{Artificial intelligence and statistics}. PMLR.

\bibitem[{Min et~al.(2020)Min, Yao, Xie, Wang, Zha, and Zhang}]{min2020domain}
Min, S.; Yao, H.; Xie, H.; Wang, C.; Zha, Z.-J.; and Zhang, Y. 2020.
\newblock Domain-aware visual bias eliminating for generalized zero-shot
  learning.
\newblock In \emph{CVPR}.

\bibitem[{Norouzi et~al.(2014)Norouzi, Mikolov, Bengio, Singer, Shlens, Frome,
  Corrado, and Dean}]{norouzi2013zero}
Norouzi, M.; Mikolov, T.; Bengio, S.; Singer, Y.; Shlens, J.; Frome, A.;
  Corrado, G.~S.; and Dean, J. 2014.
\newblock Zero-shot learning by convex combination of semantic embeddings.
\newblock In \emph{ICLR}.

\bibitem[{Pambala, Dutta, and Biswas(2020)}]{pambala2020generative}
Pambala, A.; Dutta, T.; and Biswas, S. 2020.
\newblock Generative model with semantic embedding and integrated classifier
  for generalized zero-shot learning.
\newblock In \emph{WACV}.

\bibitem[{Qin et~al.(2016)Qin, Wang, Liu, Chen, and Shao}]{qin2016beyond}
Qin, J.; Wang, Y.; Liu, L.; Chen, J.; and Shao, L. 2016.
\newblock Beyond Semantic Attributes: Discrete Latent Attributes Learning for
  Zero-Shot Recognition.
\newblock \emph{IEEE SPL}, 23(11): 1667--1671.

\bibitem[{Romera-Paredes and Torr(2015)}]{romera2015embarrassingly}
Romera-Paredes, B.; and Torr, P. 2015.
\newblock An embarrassingly simple approach to zero-shot learning.
\newblock In \emph{ICML}.

\bibitem[{Sariyildiz and Cinbis(2019)}]{sariyildiz2019gradient}
Sariyildiz, M.~B.; and Cinbis, R.~G. 2019.
\newblock Gradient matching generative networks for zero-shot learning.
\newblock In \emph{CVPR}.

\bibitem[{Smith et~al.(2021)Smith, Hsu, Balloch, Shen, Jin, and
  Kira}]{smith2021always}
Smith, J.; Hsu, Y.-C.; Balloch, J.; Shen, Y.; Jin, H.; and Kira, Z. 2021.
\newblock Always Be Dreaming: A New Approach for Data-Free Class-Incremental
  Learning.
\newblock \emph{arXiv preprint arXiv:2106.09701}.

\bibitem[{Song et~al.(2018)Song, Shen, Yang, Liu, and
  Song}]{song2018transductive}
Song, J.; Shen, C.; Yang, Y.; Liu, Y.; and Song, M. 2018.
\newblock Transductive unbiased embedding for zero-shot learning.
\newblock In \emph{CVPR}, 1024--1033.

\bibitem[{Sudlow et~al.(2015)Sudlow, Gallacher, Allen, Beral, Burton, Danesh,
  Downey, Elliott, Green, Landray et~al.}]{ukbiobank}
Sudlow, C.; Gallacher, J.; Allen, N.; Beral, V.; Burton, P.; Danesh, J.;
  Downey, P.; Elliott, P.; Green, J.; Landray, M.; et~al. 2015.
\newblock UK biobank: an open access resource for identifying the causes of a
  wide range of complex diseases of middle and old age.
\newblock \emph{PLoS medicine}, 12(3): e1001779.

\bibitem[{Verma, Arora, and Mishra(2018)}]{verma2017generalized}
Verma, V.~K.; Arora, G.; and Mishra. 2018.
\newblock Generalized zero-shot learning via synthesized examples.
\newblock In \emph{CVPR}.

\bibitem[{Voigt and Von~dem Bussche(2017)}]{voigt2017eu}
Voigt, P.; and Von~dem Bussche, A. 2017.
\newblock The eu general data protection regulation (gdpr).
\newblock \emph{A Practical Guide, 1st Ed., Cham: Springer International
  Publishing}, 10: 3152676.

\bibitem[{Vyas, Venkateswara, and Panchanathan(2020)}]{vyas2020leveraging}
Vyas, M.~R.; Venkateswara, H.; and Panchanathan, S. 2020.
\newblock Leveraging seen and unseen semantic relationships for generative
  zero-shot learning.
\newblock In \emph{ECCV}.

\bibitem[{Xian et~al.(2017)Xian, Lampert, Schiele, and Akata}]{xian2017zero}
Xian, Y.; Lampert, C.~H.; Schiele, B.; and Akata, Z. 2017.
\newblock Zero-shot learning-A comprehensive evaluation of the good, the bad
  and the ugly.
\newblock In \emph{CVPR}.

\bibitem[{Xian et~al.(2018)Xian, Lorenz, Schiele, and Akata}]{xian2018feature}
Xian, Y.; Lorenz, T.; Schiele, B.; and Akata, Z. 2018.
\newblock Feature generating networks for zero-shot learning.
\newblock In \emph{CVPR}.

\bibitem[{Xu et~al.(2015)Xu, Wang, Chen, and Li}]{Xu2015Empirical}
Xu, B.; Wang, N.; Chen, T.; and Li, M. 2015.
\newblock Empirical Evaluation of Rectified Activations in Convolutional
  Network.
\newblock In \emph{Proc. ICML Deep Learning Workshop}.

\bibitem[{Xu, Liu, and Loy(2020)}]{xu2020computation}
Xu, G.; Liu, Z.; and Loy, C.~C. 2020.
\newblock Computation-Efficient Knowledge Distillation via Uncertainty-Aware
  Mixup.
\newblock \emph{arXiv preprint arXiv:2012.09413}.

\bibitem[{Ye and Guo(2019)}]{ye2019progressive}
Ye, M.; and Guo, Y. 2019.
\newblock Progressive ensemble networks for zero-shot recognition.
\newblock In \emph{CVPR}.

\bibitem[{Yin et~al.(2021)Yin, Mallya, Vahdat, Alvarez, Kautz, and
  Molchanov}]{yin2021see}
Yin, H.; Mallya, A.; Vahdat, A.; Alvarez, J.~M.; Kautz, J.; and Molchanov, P.
  2021.
\newblock See through Gradients: Image Batch Recovery via GradInversion.
\newblock In \emph{Proceedings of the IEEE/CVF Conference on Computer Vision
  and Pattern Recognition}, 16337--16346.

\bibitem[{Yin et~al.(2020)Yin, Molchanov, Alvarez, Li, Mallya, Hoiem, Jha, and
  Kautz}]{yin2020dreaming}
Yin, H.; Molchanov, P.; Alvarez, J.~M.; Li, Z.; Mallya, A.; Hoiem, D.; Jha,
  N.~K.; and Kautz, J. 2020.
\newblock Dreaming to distill: Data-free knowledge transfer via deepinversion.
\newblock In \emph{Proceedings of the IEEE/CVF Conference on Computer Vision
  and Pattern Recognition}, 8715--8724.

\bibitem[{Zhang et~al.(2020{\natexlab{a}})Zhang, Liu, Long, Zhang, and
  Shao}]{zhang2020deep}
Zhang, H.; Liu, L.; Long, Y.; Zhang, Z.; and Shao, L. 2020{\natexlab{a}}.
\newblock Deep transductive network for generalized zero shot learning.
\newblock \emph{Pattern Recognition}, 105: 107370.

\bibitem[{Zhang et~al.(2018)Zhang, Long, Guan, and Shao}]{zhang2018triple}
Zhang, H.; Long, Y.; Guan, Y.; and Shao, L. 2018.
\newblock Triple verification network for generalized zero-shot learning.
\newblock \emph{IEEE Transactions on Image Processing}, 28(1): 506--517.

\bibitem[{Zhang et~al.(2020{\natexlab{b}})Zhang, Wang, Liu, Shen, Wei, Zhang,
  and Van Den~Hengel}]{zhang2020towards}
Zhang, L.; Wang, P.; Liu, L.; Shen, C.; Wei, W.; Zhang, Y.; and Van Den~Hengel,
  A. 2020{\natexlab{b}}.
\newblock Towards effective deep embedding for zero-shot learning.
\newblock \emph{IEEE Transactions on Circuits and Systems for Video
  Technology}, 30(9): 2843--2852.

\bibitem[{Zhang, Xiang, and Gong(2017)}]{zhang2017learning}
Zhang, L.; Xiang, T.; and Gong, S. 2017.
\newblock Learning a deep embedding model for zero-shot learning.
\newblock In \emph{CVPR}.

\bibitem[{Zhang and Saligrama(2016)}]{zhang2016zero}
Zhang, Z.; and Saligrama, V. 2016.
\newblock Zero-shot recognition via structured prediction.
\newblock In \emph{ECCV}.

\bibitem[{Zhu, Liu, and Han(2019)}]{zhu2019deep}
Zhu, L.; Liu, Z.; and Han, S. 2019.
\newblock Deep Leakage from Gradients.
\newblock \emph{Advances in NeurIPS}, 32: 14774--14784.

\bibitem[{Zhu(2015)}]{zhu2015machine}
Zhu, X. 2015.
\newblock Machine teaching: An inverse problem to machine learning and an
  approach toward optimal education.
\newblock In \emph{Proceedings of the AAAI Conference on Artificial
  Intelligence}, volume~29.

\end{thebibliography}

\end{document}